\documentclass{article}

\usepackage[final, nonatbib]{nips_2018}

\usepackage[utf8]{inputenc} 
\usepackage[T1]{fontenc}    
\usepackage{hyperref}       
\usepackage{url}            
\usepackage{booktabs}       
\usepackage{amsfonts}       
\usepackage{nicefrac}       
\usepackage{microtype}      

\usepackage{color}
\usepackage{algorithm}
\usepackage[noend]{algpseudocode}
\usepackage{mathrsfs}
\usepackage{dsfont}
\usepackage{lmodern}
\usepackage{array}
\usepackage{bbm}
\usepackage{amsmath}
\usepackage{amssymb}
\usepackage[mathscr]{eucal}
\usepackage{graphicx}
\usepackage{mathrsfs}
\usepackage{psfrag}
\usepackage{color}
\usepackage{here}
\usepackage{wasysym}
\usepackage{enumitem}
\usepackage[dvipsnames]{xcolor}
\usepackage{caption}
\usepackage{subcaption}
\usepackage{amsthm}
\usepackage{lipsum}
\usepackage{todonotes}
\usepackage{tabulary}
\usepackage{pifont}

\newtheorem{lemma} {Lemma}

\newcommand{\Exp}{\mathds{E}}

\newcommand{\Prob}{\mathds{P}}
\newcommand{\Real}{\mathds{R}}
\newcommand{\Nat}{\mathbb{N}}
\newcommand{\Ind}{\mathds{1}}

\DeclareMathOperator*{\argmin}{arg\,min}

\newcommand{\Xc}{\mathcal{X}}
\newcommand{\Yc}{\mathcal{Y}}
\newcommand{\Pc}{\mathcal{P}}

\newcommand{\Rc}{\mathcal{R}}
\newcommand{\Sc}{\mathcal{S}}
\newcommand{\Ac}{\mathcal{A}}

\newcommand{\Dc}{\mathcal{D}}

\newcommand{\Hc}{\mathcal{H}}
\newcommand{\Lc}{\mathcal{L}}

\newcommand{\ass}{\hspace{-1mm} = \hspace{-0.5mm} \mathbf{\cdot} \hspace{0.5mm}}
\newcommand{\tick}{\textcolor{ForestGreen}{\ding{51}}}
\newcommand{\ok}{\textcolor{Dandelion}{\ding{108}}}
\newcommand{\cross}{\textcolor{BrickRed}{\ding{55}}}

\definecolor{ian_highlight}{RGB}{100, 2, 2}

\errorcontextlines\maxdimen

\makeatletter
    \newcommand*{\algrule}[1][\algorithmicindent]{\makebox[#1][l]{\hspace*{.5em}\thealgruleextra\vrule height \thealgruleheight depth \thealgruledepth}}%
\newcommand*{\thealgruleextra}{}
\newcommand*{\thealgruleheight}{.75\baselineskip}
\newcommand*{\thealgruledepth}{.25\baselineskip}

\newcount\ALG@printindent@tempcnta
\def\ALG@printindent{%
    \ifnum \theALG@nested>0
        \ifx\ALG@text\ALG@x@notext
        \else
            \unskip
            \addvspace{-1pt}
            \ALG@printindent@tempcnta=1
            \loop
                \algrule[\csname ALG@ind@\the\ALG@printindent@tempcnta\endcsname]%
                \advance \ALG@printindent@tempcnta 1
            \ifnum \ALG@printindent@tempcnta<\numexpr\theALG@nested+1\relax
            \repeat
        \fi
    \fi
    }%
\usepackage{etoolbox}
\patchcmd{\ALG@doentity}{\noindent\hskip\ALG@tlm}{\ALG@printindent}{}{\errmessage{failed to patch}}
\makeatother

\newbox\statebox
\newcommand{\myState}[1]{%
    \setbox\statebox=\vbox{#1}%
    \edef\thealgruleheight{\dimexpr \the\ht\statebox+1pt\relax}%
    \edef\thealgruledepth{\dimexpr \the\dp\statebox+1pt\relax}%
    \ifdim\thealgruleheight<.75\baselineskip
        \def\thealgruleheight{\dimexpr .75\baselineskip+1pt\relax}%
    \fi
    \ifdim\thealgruledepth<.25\baselineskip
        \def\thealgruledepth{\dimexpr .25\baselineskip+1pt\relax}%
    \fi
    \State #1%
    \def\thealgruleheight{\dimexpr .75\baselineskip+1pt\relax}%
    \def\thealgruledepth{\dimexpr .25\baselineskip+1pt\relax}%
}

\title{Randomized Prior Functions \\ for Deep Reinforcement Learning}

\author{
  Ian Osband \\
  DeepMind \\
  \texttt{iosband@google.com} \\
  \And
  John Aslanides \\
  DeepMind \\
  \texttt{jaslanides@google.com} \\
  \And
  Albin Cassirer \\
  DeepMind \\
  \texttt{cassirer@google.com} \\
}

\begin{document}

\maketitle

\begin{abstract}

Dealing with uncertainty is essential for efficient reinforcement learning.
There is a growing literature on uncertainty estimation for deep learning from fixed datasets, but many of the most popular approaches are poorly-suited to sequential decision problems.
Other methods, such as bootstrap sampling, have no mechanism for uncertainty that does not come from the observed data.
We highlight why this can be a crucial shortcoming and propose a simple remedy through addition of a randomized untrainable `prior' network to each ensemble member.
We prove that this approach is efficient with linear representations, provide simple illustrations of its efficacy with nonlinear representations and show that this approach scales to large-scale problems far better than previous attempts.

\end{abstract}

\section{Introduction}
\label{sec:intro}

Deep learning methods have emerged as the state of the art approach for many challenging problems \cite{krizhevsky2012imagenet,van2016wavenet}.
This is due to the statistical flexibility and computational scalability of large and deep neural networks, which allows them to harness the information in large and rich datasets.
Deep reinforcement learning combines deep learning with sequential decision making under uncertainty.
Here an agent takes actions inside an environment in order to maximize some cumulative reward \cite{Sutton2017}.
This combination of deep learning with reinforcement learning (RL) has proved remarkably successful \cite{tesauro1995temporal,mnih2015human,silver2016alphago}.

At the same time, elementary decision theory shows that the only admissible decision rules are Bayesian \cite{cox1979theoretical,wald1992statistical}.
Colloquially, this means that any decision rule that is not Bayesian can be improved (or even exploited) by some Bayesian alternative \cite{de1937prevision}.
Despite this fact, the majority of deep learning research has evolved outside of Bayesian (or even statistical) analysis \cite{rumelhart1985learning,lecun2015deep}.
This disconnect extends to deep RL, where the majority of state of the art algorithms have no concept of uncertainty \cite{mnih2015human,mnih2016asynchronous} and can fail spectacularly even in simple problems where success requires its consideration \cite{mihatsch2002risk,osband2016}.

There is a long history of research in Bayesian neural networks that never quite became mainstream practice \cite{mackay1992practical,neal2012bayesian}.
Recently, Bayesian deep learning has experienced a resurgence of interest with a myriad of approaches for uncertainty quantification in fixed datasets and also sequential decision problems \cite{kingma2013auto,blundell2015weight,Gal2016Dropout,osband2016deep}.
In this paper we highlight the surprising fact that many of these well-cited and popular methods for uncertainty estimation in deep learning can be poor choices for sequential decision problems.
We show that this disconnect is more than a technical detail, but a serious shortcoming that can lead to arbitrarily poor performance.
We support our claims by a series of simple lemmas for simple environments, together with experimental evidence in more complex settings.

Our approach builds on an alternative method for uncertainty in deep RL inspired by the statistical bootstrap \cite{efron1982jackknife}.
This approach trains an ensemble of models, each on perturbed versions of the data.
The resulting distribution \textit{of the ensemble} is used to approximate the uncertainty in the estimate \cite{osband2016deep}.
Although typically regarded as a frequentist method, bootstrapping gives near-optimal convergence rates when used as an approximate Bayesian posterior \cite{fushiki2005nonparametric,fushiki2005bootstrap}.
However, these ensemble-based approaches to uncertainty quantification approximate a `posterior' without an effective methodology to inject a `prior'.
This can be a crucial shortcoming in sequential decision problems.

In this paper, we present a simple modification where each member of the ensemble is initialized together with a random but fixed \textit{prior function}.
Predictions in each ensemble member are then taken as the sum of the trainable neural network and its prior function.
Learning/optimization is performed so that this sum (network plus prior) minimizes training loss.
Therefore, with sufficient network capacity and optimization, the ensemble members will agree at observed data.
However, in regions of the space with little or no training data, their predictions will be determined by the generalization of their networks and priors.
Surprisingly, we show that this approach is equivalent to exact Bayesian inference for the special case of Gaussian linear models.
Following on from this `sanity check', we present a series of simple experiments designed to extend this intuition to deep learning.
We show that many of the most popular approaches for uncertainty estimation in deep RL do \textit{not} pass these sanity checks, and crystallize these shortcomings in a series of lemmas and small examples.
We demonstrate that our simple modification can facilitate aspiration in difficult tasks where previous approaches for deep RL fail.
We believe that this work presents a simple and practical approach to encoding prior knowledge with deep reinforcement learning.

\section{Why do we need a `prior' mechanism for deep RL?}
\label{sec:why_prior}

We model the environment as a Markov decision process $M = (\Sc, \Ac, R, P)$ \cite{Bertsekas1996}.
Here $\Sc$ is the state space and $\Ac$ is the action space.
At each time step $t$, the agent observes state $s_t \in \Sc$, takes action $a_t \in \Ac$, receives reward $r_t \sim R(s_t,a_t)$ and transitions to $s_{t+1} \sim P(s_t,a_t)$.
A policy $\pi : \Sc \rightarrow \Ac$ maps states to actions and let $\Hc_t$ denote the history of observations before time $t$.
An RL algorithm maps $\Hc_t$ to a distribution over policies; we assess its quality through the cumulative reward over unknown environments.
To perform well, an RL algorithm must learn to optimize its actions, combining both learning and control \cite{Sutton2017}.
A `deep' RL algorithm uses neural networks for nonlinear function approximation \cite{lecun2015deep,mnih2015human}.

The scale and scope of problems that might be approached through deep RL is vast, but there are three key aspects an efficient (and general) agent must address \cite{Sutton2017}:
\begin{enumerate}[noitemsep, nolistsep]
  \item {\bf Generalization}: be able to learn from data it collects.
  \item {\bf Exploration}: prioritize the best experiences to learn from.
  \item {\bf Long-term consequences}: consider external effects beyond a single time step.
\end{enumerate}
In this paper we focus on the importance of some form of `prior' mechanism for efficient exploration.
As a motivating example we consider a sparse reward task where random actions are very unlikely to ever see a reward.
If an agent has never seen a reward then it is essential that some other form of aspiration, motivation, drive or curiosity direct its learning.
We call this type of drive a `prior' effect, since it does not come from the observed data, but are ambivalent as to whether this effect is philosophically `Bayesian'.
Agents that do not have this prior drive will be left floundering aimlessly and thus may require exponentially large amounts of data in order to learn even simple problems \cite{kearns2002near}.

To solve a specific task, it can be possible to attain superhuman performance without significant prior mechanism \cite{mnih2015human,mnih2016asynchronous}.
However, if our goal is artificial \textit{general} intelligence, then it is disconcerting that our best agents can perform very poorly even in simple problems \cite{legg2007collection,mania2018simple}.
One potentially general approach to decision making is given by the Thompson sampling heuristic\footnote{This heuristic is general in the sense that Thompson sampling can be theoretically justified in many of the domains where these other approaches fail \cite{agrawal2012analysis,Osband2013,leike2016thompson,russo2017tutorial}.}: `randomly take action according to the probability you believe it is the optimal action' \cite{Thompson1933}.
In recent years there have been several attempts to apply this heuristic to deep reinforcement learning, each attaining significant outperformance over deep RL baselines on certain tasks \cite{Gal2016Dropout,osband2016deep,lipton2016efficient,blundell2015weight,fortunato2017noisy}.
In this section we outline crucial shortcomings for the most popular existing approaches to posterior approximation; these outlines will be brief, but more detail can be found in Appendix \ref{app:existing_bad}.
These shortcomings set the scene for Section \ref{sec:prior}, where we introduce a simple and practical alternative that passes each of our simple sanity checks: bootstrapped ensembles with randomized prior functions.
In Section \ref{sec:rl} we demonstrate that this approach scales gracefully to complex domains with deep RL.

\subsection{Dropout as posterior approximation}
\label{sec:dropout}

One of the most popular modern approaches to regularization in deep learning is dropout sampling \cite{srivastava2014dropout}.
During training, dropout applies an independent random Bernoulli mask to the activations and thus guards against excessive co-adaptation of weights.
Recent work has sought to understand dropout through a Bayesian lens, highlighting the connection to variational inference and arguing that the resultant dropout distribution approximates a Bayesian posterior \cite{Gal2016Dropout}.
This narrative has proved popular despite the fact that dropout distribution can be a poor approximation to most reasonable Bayesian posteriors \cite{gal2016improving,osband2016risk}:

\begin{lemma}[Dropout distribution does not concentrate with observed data]
\label{thm:dropout}
\hspace{0.000001mm} \newline
{\medmuskip=0mu
\thinmuskip=0mu
\thickmuskip=1mu
Consider any loss function $\Lc$, regularizer $\Rc$ and data $\Dc = \{(x,y)\}$.
Let $\theta$ be parameters of any neural network architecture $f$ trained with dropout rate $p \in (0, 1)$ and dropout masks $W$,}
\begin{equation}
\label{eq:dropout}
  \theta^*_p \in \argmin_\theta \Exp_{W \sim {\rm Ber}(p), (x,y) \sim \Dc}\left[ \Lc(x,y \mid \theta, W) + \Rc(\theta) \right].
\end{equation}
Then the dropout distribution $f_{\theta^*_p, W}$ is invariant to duplicates of the dataset $\Dc$.

\end{lemma}

Lemma \ref{thm:dropout} is somewhat contrived, but highlights a clear shortcoming of dropout as posterior sampling: the dropout rate does not depend on the data.
Lemma \ref{thm:dropout} means no agent employing dropout for posterior approximation can tell the difference between observing a set of data once and observing it $N \gg 1$ times.
This can lead to arbitrarily poor decision making, even when combined with an efficient strategy for exploration \cite{osband2016}.

\subsection{Variational inference and Bellman error}
\label{sec:variational_inference}
\vspace{-1mm}

Dropout as posterior is motivated by its connection to variational inference (VI) \cite{Gal2016Dropout}, and recent work to address Lemma \ref{thm:dropout} improves the quality of this variational approximation by tuning the dropout rate from data \cite{gal2017concrete}.\footnote{Concrete dropout assymptotically improves the quality of the variational approximation, but provides no guarantees on its rate of convergence or error relative to exact Bayesian inference \cite{gal2017concrete}.}
However, there is a deeper problem to this line of research that is common across many works in this field: even given access to an oracle method for \textit{exact} inference, applying independent inference to the Bellman error does not propagate uncertainty correctly for the value function as a whole \cite{o2017uncertainty}.
To estimate the uncertainty in $Q$ from the Bellman equation 
{\medmuskip=0mu
\thinmuskip=0mu
\thickmuskip=1mu
$Q(s_t,a_t) = \Exp[r_{t+1} + \gamma \max_\alpha Q(s_{t+1}, \alpha)]$
} it is crucial that the two sides of this equation are not independent random variables.
Ignoring this dependence can lead to very bad estimates, even with exact inference.
\vspace{1mm}

\begin{lemma}[Independent VI on Bellman error does not propagate uncertainty]
\label{thm:variational}
\hspace{0.000001mm} \newline
{\medmuskip=0mu
\thinmuskip=0mu
\thickmuskip=1mu
Let $Y \sim N(\mu_Y, \sigma_Y^2)$ be a target value. If we train $X \sim N(\mu, \sigma^2)$ according to the squared error}
\begin{equation}
\label{eq:vi_bellman}
  \mu^*, \sigma^* \in \argmin_{\mu, \sigma} \Exp \left[ (X - Y)^2 \right]
  \quad \text{ for } X, Y \text{independent},
\end{equation}
then the solution $\mu^*=\mu_Y, \sigma^*=0$ propagates zero uncertainty from $Y$ to $X$.
\end{lemma}

{\medmuskip=0mu
\thinmuskip=0mu
\thickmuskip=1mu
To understand the significance of Lemma \ref{thm:variational}, imagine a deterministic system that transitions from $s_1$ to $s_2$ without reward.
Suppose an agent is able to correctly quantify their posterior uncertainty for the value $V(s_2) = Y \sim N(\mu_Y, \sigma_Y^2)$.
Training $V(s_1) = X$ according to \eqref{eq:vi_bellman} will lead to zero uncertainty estimates at $s_1$, when in fact $V(s_1) \sim N(\mu_Y, \sigma_Y^2)$.
This observation may appear simplistic, and may not say much more than `do not use the squared loss' for VI in this setting.
However, despite this clear failing \eqref{eq:vi_bellman} is precisely the loss used by the majority of approaches to VI for  RL \cite{fortunato2017noisy,lipton2016efficient,tang2017variational,touati2018randomized,Gal2016Dropout}.
Note that this failure occurs even without decision making, function approximation and even when the true posterior lies within the variational class.
}

\subsection{`Distributional reinforcement learning'}
\label{sec:distributional}
\vspace{-1mm}

The key ingredient for a Bayesian formulation for sequential decision making is to consider beliefs not simply as a point estimate, but as a \textit{distribution}.
Recently an approach called `distributional RL' has shown great success in improving stability and performance in deep RL benchmark algorithms \cite{bellemare2017distributional}.
Despite the name, these two ideas are quite distinct.
`Distributional RL' replaces a scalar estimate for the value function by a distribution that is trained to minimize a loss against the distribution of data it observes.
This distribution of observed data is an orthogonal concept to that of Bayesian uncertainty.

\vspace{-1mm}
\begin{figure}[h!]
\centering
\begin{subfigure}{0.48\linewidth}
    \centering
    \includegraphics[width=\linewidth]{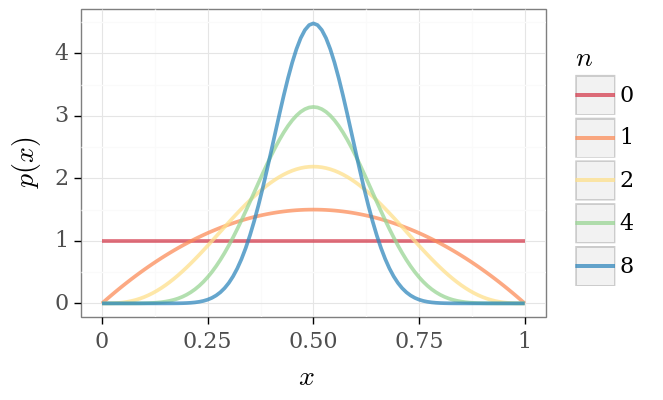}
    \vspace{-6mm}
    \caption{Posterior beliefs concentrate around $p=0.5$.}
    \label{fig:dist-bayes-conc}
\end{subfigure}
\quad
{\medmuskip=0mu
\thinmuskip=0mu
\thickmuskip=1mu
\begin{subfigure}{0.48\linewidth}
    \centering
    \includegraphics[width=\linewidth]{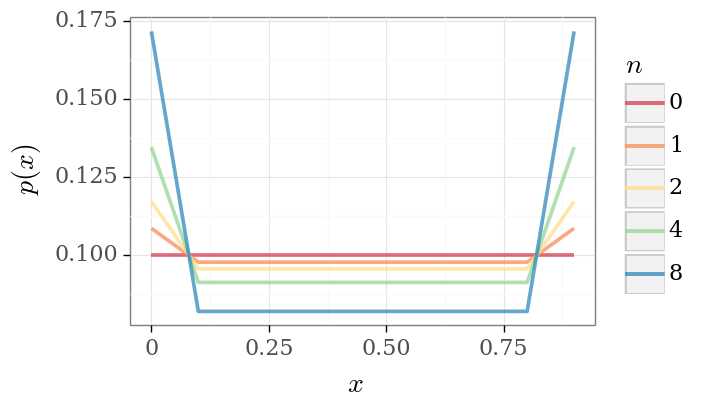}
    \vspace{-4mm}
    \caption{`Distributional' tends to mass at $0$ and $1$.}
    \label{fig:dist-dist-conc}
\end{subfigure}
}
\centering
\vspace{-2mm}
\caption{Output distribution after observing $n$ heads and $n$ tails of a coin.}
\label{fig:distributional}
\end{figure}
\vspace{-1mm}

Figure \ref{fig:distributional} presents an illustration of these two distributions after observing flips of a coin.
As more data is gathered the posterior distribution concentrates around the mean whereas the `distributional RL' estimate approaches that of the generating Bernoulli.
Although both approaches might reasonably claim a `distributional perspective' on RL, these two distributions have orthogonal natures and behave quite differently.
Conflating one for the other can lead to arbitrarily poor decisions; it is the uncertainty in beliefs (`epistemic'), not the distributional noise (`aleatoric') that is important for exploration \cite{kearns2002near}.

\subsection{Count-based uncertainty estimates}
\label{sec:count}
\vspace{-1mm}

Another popular method for incentivizing exploration is with a density model or `pseudocount' \cite{bellemare2016countbased}.
Inspired by the analysis of tabular systems, these models assign a bonus to states and actions that have been visited infrequently according to a density model.
This method can perform well, but only when the generalization of the density model is aligned with the task objective.
Crucially, this generalization is not learned from the task \cite{ostrovski2017countbased}.

\begin{figure}[h!]
\centering
\begin{subfigure}{0.48\linewidth}
    \centering
    \includegraphics[width=\linewidth]{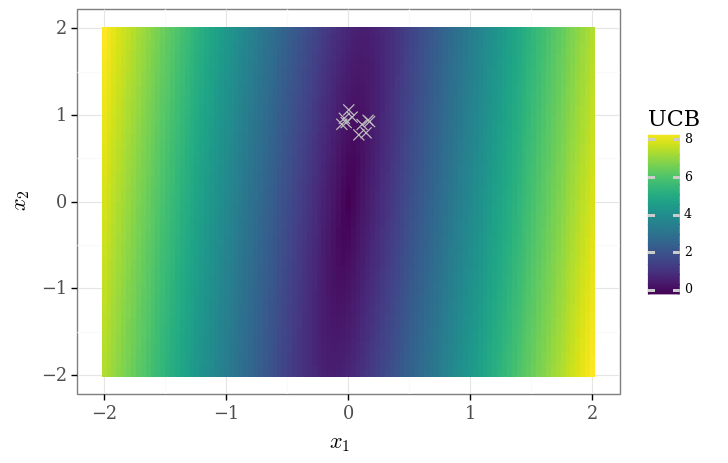}
    \vspace{-2mm}
    \caption{Uncertainty bonus from posterior over $x^T \theta^*$.}
    \label{fig:confidence_correct}
\end{subfigure}
\quad
{\medmuskip=0mu
\thinmuskip=0mu
\thickmuskip=1mu
\begin{subfigure}{0.48\linewidth}
    \centering
    \includegraphics[width=\linewidth]{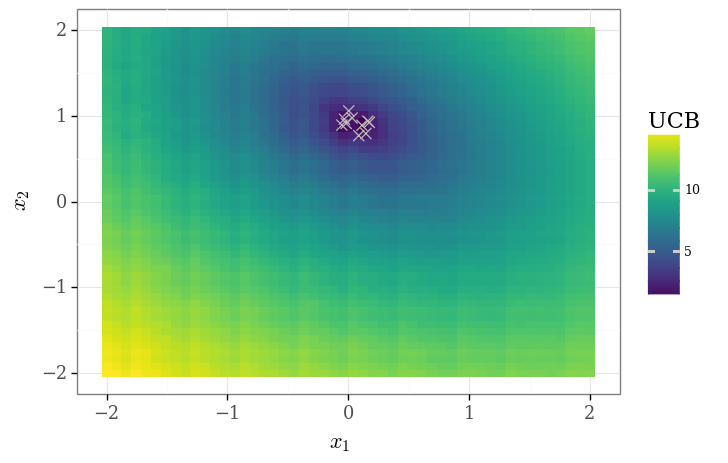}
    \vspace{-2mm}
    \caption{Bonus from Gaussian pseudocount $p(x)$.}
    \label{fig:confidence_count}
\end{subfigure}
}
\centering
\vspace{-2mm}
\caption{Count-based uncertainty leads to a poorly aligned bonus even in a linear system.}
\label{fig:count}
\end{figure}

Even with an optimal state representation and density, a count-based bonus on states can be poorly suited for efficient exploration.
Consider a linear bandit with reward $r_t(x_t) = x_t^T \theta^* + \epsilon_t$ for some $\theta^* \in \Real^d$ and $\epsilon_t \sim N(0,1)$ \cite{RusmevichientongT2010}.
Figure \ref{fig:count} compares the uncertainty in the expected reward $\Exp[x^T \theta^*]$ with that obtained by density estimation on the observed $x_t$.
A bonus based upon the state density does not correlate with the \textit{uncertainty} over the unknown optimal action.
This disconnect can lead to arbitrarily poor decisions \cite{osband2017deep}.

\vspace{-2mm}
\vspace{-2mm}

\section{Randomized prior functions for deep ensembles}
\label{sec:prior}

Section \ref{sec:why_prior} motivates the need for effective uncertainty estimates in deep RL.
We note that crucial failure cases of several popular approaches can arise even with simple linear models.
As a result, we take a moment to review the setting of Bayesian linear regression.
Let $\theta \in \Real^d$ with prior $N(\overline{\theta}, \lambda I)$ and data $\Dc = \left\{(x_i, y_i)\right\}_{i=1}^{n}$ for $x_i \in \Real^d$ and $y_i = \theta^T x_i + \epsilon_i$ with $\epsilon_i \sim N(0, \sigma^2)$ iid.
Then, conditioned on $\Dc$, the posterior for $\theta$ is Gaussian:
{\medmuskip=0mu
\thinmuskip=0mu
\thickmuskip=1mu
\begin{equation}
  \label{eq:blr}
  \Exp\left[\theta \mid \Dc \right] =
  \left( \frac{1}{\sigma^2} X^T X + \frac{1}{\lambda} I \right)^{-1} \left(\frac{1}{\sigma^2} X^T y + \frac{1}{\lambda}\overline{\theta} \right),
  \quad
  {\rm Cov}\left[\theta \mid \Dc\right] = \left( \frac{1}{\sigma^2} X^T X + \frac{1}{\lambda} I \right)^{-1}.
\end{equation}
}
\hspace{-2mm} Equation \eqref{eq:blr} relies on Gaussian conjugacy and linear models, which cannot easily be extended to deep neural networks.
The following result shows that we can replace this analytical result with a simple computational procedure.

\begin{lemma}[Computational generation of posterior samples]
\label{thm:correctness}
\hspace{0.000001mm} \newline
Let $f_\theta(x) = x^T \theta$, $\tilde{y}_i \sim N(y_i, \sigma^2)$ and $\tilde{\theta} \sim N(\overline{\theta}, \lambda I)$.
Then the either of the following optimization problems generate a sample $\theta \mid \Dc$ according to \eqref{eq:blr}:
{
\begin{eqnarray}
\label{eq:l2_weight_blr}
  \argmin_\theta \sum_{i=1}^n \|\tilde{y}_i - f_{\theta}(x_i) \|^2_2
    + \frac{\sigma^2}{\lambda} \|\tilde{\theta} - \theta\|^2_2, \\
    \label{eq:l2_weight_blr_2}
  \tilde{\theta} + \argmin_\theta \sum_{i=1}^n \|\tilde{y}_i - \left(f_{\tilde{\theta}} + f_{\theta}\right)(x_i) \|^2_2
    + \frac{\sigma^2}{\lambda} \|\theta\|^2_2.
\end{eqnarray}}
\vspace{-5mm}
\proof
To prove \eqref{eq:l2_weight_blr} note that the solution is Gaussian and then match moments; equation \eqref{eq:l2_weight_blr_2} then follows by relabeling \cite{osband2017deep}.
\qed
\end{lemma}

Lemma \ref{thm:correctness} is revealing since it allows us to view Bayesian regression through a purely computational lens: `generate posterior samples by training on noisy versions of the data, together with some random regularization'.
Even for nonlinear $f_\theta$, we can still compute \eqref{eq:l2_weight_blr} or \eqref{eq:l2_weight_blr_2}.
Although the resultant $f_\theta$ will no longer be an exact posterior, at least it passes the `sanity check' in this simple linear setting (unlike the approaches of Section \ref{sec:why_prior}).
We argue this method is quite intuitive: the perturbed data $\tilde{\Dc} = \{(x_i, \tilde{y}_i)\}_{i=1}^n$ is generated according to the estimated noise process $\epsilon_t$ and the sample $\tilde{\theta}$ is drawn from prior beliefs.
Intuitively \eqref{eq:l2_weight_blr} says to fit to $\tilde{\Dc}$ and regularize weights to a prior sample of weights $\tilde{\theta}$; \eqref{eq:l2_weight_blr_2} says to generate a prior \textit{function} $f_{\tilde{\theta}}$ and then fit an additive term to noisy data $\tilde{\Dc}$ with regularized complexity.

This paper explores the performance of each of these methods for uncertainty estimation with deep learning.
We find empirical support that method \eqref{eq:l2_weight_blr_2} coupled with a \textit{randomized prior function} significantly outperforms ensemble-based approaches without prior mechanism.
We also find that \eqref{eq:l2_weight_blr_2} significantly outperforms \eqref{eq:l2_weight_blr} in deep RL.
We suggest a major factor in this comes down to the huge dimensionality of neural network weights, whereas the output policy or value is typically far smaller.
In this case, it makes sense to enforce prior beliefs in the low dimensional space.
Further, the initialization of neural network weights plays an important role in their generalization properties and optimization via stochastic gradient descent (SGD) \cite{glorot2010understanding,maennel2018gradient}.
As such, \eqref{eq:l2_weight_blr_2} may help to decouple the dual roles of initial weights as both `prior' and training initializer.
Algorithm \ref{alg:randomized_prior} describes our approach applied to modern deep learning architectures.

\begin{algorithm}[h!]
{\small
\caption{Randomized prior functions for ensemble posterior.}
\label{alg:randomized_prior}
\begin{algorithmic}[1]
{\medmuskip=0mu
\thinmuskip=0mu
\thickmuskip=1mu
  \Require{Data $\Dc \subseteq \{(x, y) | x \in \Xc, y \in \Yc\}$, loss function $\Lc$, neural model $f_\theta : \Xc \rightarrow \Yc$, \newline Ensemble size $K \in \Nat$, noise procedure \texttt{data\_noise}, distribution over priors $\Pc \subseteq \{ \Prob(p) \mid p : \Xc \rightarrow \Yc\}$.}}
  \For{$k = 1,..,K$}
    \State initialize $\theta_k \sim$ Glorot initialization \cite{glorot2010understanding}.
    \State form $\Dc_k = \mathtt{data\_noise}(\Dc)$ (e.g. Gaussian noise or bootstrap sampling \cite{osband2015bootstrapped}).
    \State {sample prior function $p_k \sim \Pc$.}
    \State optimize $ \nabla_{\theta \mid \theta=\theta_k} \Lc(f_{\theta} + p_k; \Dc_k)$ via ADAM \cite{kingma2014adam}.
  \EndFor
  \State {\bf return} ensemble $\{ f_{\theta_k} + p_k \}_{k=1}^K$.

\end{algorithmic}
}
\end{algorithm}

\section{Deep reinforcement learning}
\label{sec:rl}
\vspace{-2mm}

Algorithm \ref{alg:randomized_prior} might be applied to model or policy learning approaches, but this paper focuses on value learning.
We apply Algorithm \ref{alg:randomized_prior} to \textit{deep Q networks} (DQN) \cite{mnih2015human} on a series of tasks designed to require good uncertainty estimates.
We train an ensemble of $K$ networks $\{Q_k\}_{k=1}^K$ in parallel, each on a perturbed version of the observed data $\Hc_t$ and each with a distinct random, but fixed, prior function $p_k$.
Each episode, the agent selects $j \sim {\rm Unif}([1,..,K])$ and follows the greedy policy w.r.t. $Q_j$ for the duration of the episode.
This algorithm is essentially bootstrapped DQN (BootDQN) except for the addition of the prior function $p_k$ \cite{osband2016deep}.
We use the statistical bootstrap rather than Gaussian noise \eqref{eq:l2_weight_blr_2} to implement a state-specific variance \cite{fushiki2005nonparametric}.

Let $\gamma \in [0,1]$ be a discount factor that induces a time preference over future rewards.
For a neural network family $f_\theta$, prior function $p$, and data $\Dc = \{(s_t,a_t, r_t, s'_t)$ we define the $\gamma$-discounted empirical temporal difference (TD) loss,
\vspace{-1mm}
\begin{equation}
\label{eq:emp_bellman_gamma}
\vspace{-1mm}
  \Lc_\gamma(\theta ; \theta^-, p, \Dc) := 
  \sum_{t \in \Dc} \left(r_t + \gamma \max_{a' \in \Ac} \overbrace{(f_{\theta^-} + p)}^{\text{target } Q}(s'_t, a') 
  - \overbrace{(f_\theta + p)}^{\text{online } Q}(s_t, a_t) \right)^2.
\end{equation}
Using this notation, the learning update for BootDQN with prior functions is a simple application of Algorithm \ref{alg:randomized_prior}, which we outline below.
To complete the RL algorithm we implement a 50-50 $\mathtt{ensemble\_buffer}$, where each transition has a 50\% chance of being included in the replay for model $k=1,..,K$.
For a complete description of BootDQN+prior agent, see Appendix \ref{app:algorithm}.

{\small
\begin{algorithm}[!ht]
\caption{$\mathtt{learn\_bootstrapped\_dqn\_with\_prior}$}
\label{alg:learn_ensemble_rlsvi}

{\small
\begin{tabular}{lll}
\textbf{Agent:} & $\theta_1, .., \theta_K$ & trainable network parameters \\
& $p_1, .., p_K$ & fixed prior functions \\
& $\Lc_\gamma(\theta \ass ; \theta^- \ass, p \ass, \Dc \ass)$ & TD error loss function \\
& $\mathtt{ensemble\_buffer}$ & replay buffer of $K$-parallel perturbed data \\
\textbf{Updates:} & $\theta_1, .., \theta_K$ & agent value function estimate
\end{tabular}

\begin{algorithmic}[1]
\For{$k$ in $(1,\ldots,K)$}
\State Data $\Dc_k \leftarrow \mathtt{ensemble\_buffer[k].sample\_minibatch()}$

\State optimize $\nabla_{\theta \mid \theta=\theta_k} \Lc(\theta; \theta_k, p_k, \Dc_k)$ via ADAM \cite{kingma2014adam}.

\EndFor
\end{algorithmic}
}
\end{algorithm}
}

\vspace{-1mm}
\subsection{Does BootDQN+prior address the shortcomings from Section \ref{sec:why_prior}?}
\label{sec:how_does_it_solve}
\vspace{-1mm}

Algorithm \ref{alg:learn_ensemble_rlsvi} is a simple modification of vanilla Q-learning: rather than maintain a single point estimate for $Q$, we maintain $K$ estimates in parallel, and rather than regularize each estimate to a single value, each is individually regularized to a distinct random prior function.
We show that that this simple and scalable algorithm overcomes the crucial shortcomings that afflict existing methods, as outlined in Section \ref{sec:why_prior}.
\begin{itemize}[noitemsep, nolistsep, leftmargin=*]
  \item[\checkmark] {\bf Posterior concentration} (Section \ref{sec:dropout}): Prior function + noisy data means the ensemble is initially diverse, but concentrates as more data is gathered. For linear-gaussian systems this matches Bayes posterior, bootstrap offers a general, non-parametric approach \cite{efron1994introduction,fushiki2005bootstrap}.
  \item[\checkmark] {\bf Multi-step uncertainty} (Section \ref{sec:variational_inference}): Since each network $k$ trains only on its \textit{own} target value, BootDQN+prior propagates a temporally-consistent sample of $Q$-value \cite{osband2017deep}.
  \item[\checkmark] {\bf Epistemic vs aleatoric} (Section \ref{sec:distributional}): BootDQN+prior optimises the \textit{mean} TD loss \eqref{eq:emp_bellman_gamma} and does not seek to fit the noise in returns, unlike `distributional RL' \cite{c51}.
  \item[\checkmark] {\bf Task-appropriate generalization} (Section \ref{sec:count}): We explore according to our uncertainty in the value $Q$, rather than density on state. As such, our generalization naturally occurs in the space of \textit{features} relevant to the task, rather than pixels or noise \cite{bellemare2016countbased}.
  {\medmuskip=0mu
  \thinmuskip=0mu
  \thickmuskip=1mu
  \item[\checkmark] {\bf Intrinsic motivation} (comparison to BootDQN without prior): In an environment with zero rewards, a bootstrap ensemble may simply learn to predict zero for \textit{all} states.
  The prior $p_k$ can make this generalization unlikely for $Q_k$ at unseen states $\tilde{s}$ so $\Exp[\max_{\alpha}Q_k(\tilde{s}, \alpha)] > 0$; thus BootDQN+prior seeks novelty even with no observed rewards.
  }
\end{itemize}

Another source of justification comes from the observation that BootDQN+prior is an instance of \textit{randomized least-squares value iteration} (RLSVI), with regularization via `prior function' for an ensemble of neural networks.
RLSVI with linear function approximation and Gaussian noise guarantees a bound on expected regret of $\tilde{O}(\sqrt{|\Sc| |\Ac| T})$ in the tabular setting \cite{osband2017deep}.\footnote{Regret measures the shortfall in cumulative rewards compared to that of the optimal policy.}
Similarly, analysis for the bandit setting establishes that $K = \tilde{O}(|\Ac|)$ models trained online can attain similar performance to full resampling each episode \cite{lu2017ensemble}.
Our work in this paper pushes beyond the boundaries of these analyses, which are presented as `sanity checks' that our algorithm is at least sensible in simple settings, rather than a certificate of correctness for more complex ones.
The rest of this paper is dedicated to an empirical investigation of our algorithm through computational experiments.
Encouragingly, we find that many of the insights born out of simple problems extend to more complex `deep RL' settings and good evidence for the efficacy of our algorithm.



\vspace{-2mm}
\subsection{Computational experiments}
\vspace{-1mm}
Our experiments focus on a series of environments that require deep exploration together with increasing state complexity \cite{kearns2002near,osband2017deep}.
In each of our domains, random actions are very unlikely to achieve a reward and exploratory actions may even come at a cost.
Any algorithm without prior motivation will have no option but to explore randomly, or worse, eschew exploratory actions completely in the name of premature and sub-optimal exploitation.
In our experiments we focus on a \textit{tabula rasa} setting in which the prior function is drawn as a random neural network.
Although our prior distribution $\Pc$ could encode task-specific knowledge (e.g. through sampling the true $Q^*$), we leave this investigation to future work.

\vspace{-2mm}
\subsubsection{Chain environments}
\label{sec:deep_sea}
\vspace{-1mm}

{\medmuskip=0mu
\thinmuskip=0mu
\thickmuskip=1mu
We begin our experiments with a family of chain-like environments that highlight the need for deep exploration \cite{strens2000bayesian}.
The environments are indexed by problem scale $N \in \Nat$ and action mask $W \sim {\rm Ber}(0.5)^{N \times N}$, with $\Sc = \{0, 1\}^{N \times N}$ and $\Ac = \{0,1\}$.
The agent begins each episode in the upper left-most state in the grid and deterministically falls one row per time step.
The state encodes the agent's row and column as a one-hot vector $s_t \in \Sc$.
The actions $\{0, 1\}$ move the agent left or right depending on the action mask $W$ at state $s_t$, which remains fixed.
The agent incurs a cost of $0.01/N$ for moving right in all states except for the right-most, in which the reward is $1$.
The reward for action left is always zero.
An episode ends after $N$ time steps so that the optimal policy is to move right each step and receive a total return of $0.99$; all other policies receive zero or negative return.
Crucially, algorithms without deep exploration take $\Omega(2^N)$ episodes to learn the optimal policy \cite{osband2016rlsvi}.\footnote{The dashed lines indicate the $2^N$ dithering lower bound. The action mask $W$ means this cannot be solved easily by evolution or policy search evolution, unlike previous `chain' examples \cite{osband2016deep,plappert2017parameter}.}
}

\vspace{-1mm}
\begin{figure}[h!]
  \centering
  \includegraphics[width=\linewidth]{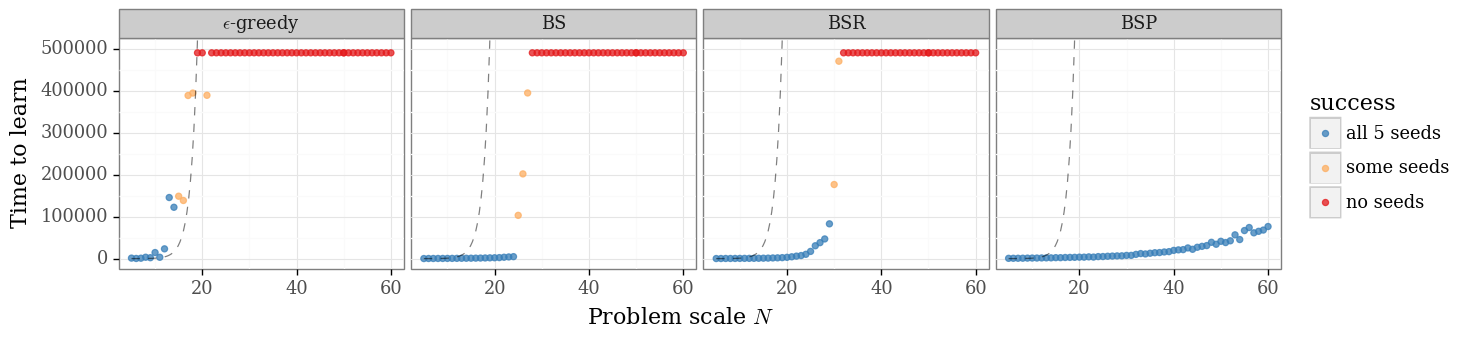}
  \vspace{-6mm}
  \caption{\small Only bootstrap with additive prior network (BSP) scales gracefully to large problems. Plotting BSP on a log-log scale suggests an empirical scaling $T_{\rm learn} = \tilde{O}(N^3)$; see Figure \ref{fig:learn_log_scale}.}
  \label{fig:deep_sea_scale}
  \vspace{-1mm}
\end{figure}

Figure \ref{fig:deep_sea_scale} presents the average time to learn for $N=5,..,60$ up to $500$K  episodes over 5 seeds and ensemble $K=20$.
We say that an agent has learned the optimal policy when the average regret per episode drops below $0.9$.
We compare three variants of BootDQN, depending on their mechanism for `prior' effects.
{\bf BS} is bootstrap without prior mechanism. {\bf BSR} is bootstrap with $l_2$ regularization on weights per \eqref{eq:l2_weight_blr}. {\bf BSP} is bootstrap with additive prior function per \eqref{eq:l2_weight_blr_2}.
In each case we initialize a random 20-unit MLP; BSR regularizes to these initial weights and BSP trains an additive network.
Although all bootstrap methods significantly outperform $\epsilon$-greedy, only BSP successfully scales to large problem sizes.

\begin{figure}[h!]
\begin{subfigure}{0.48\linewidth}
    \centering
    \includegraphics[width=\linewidth]{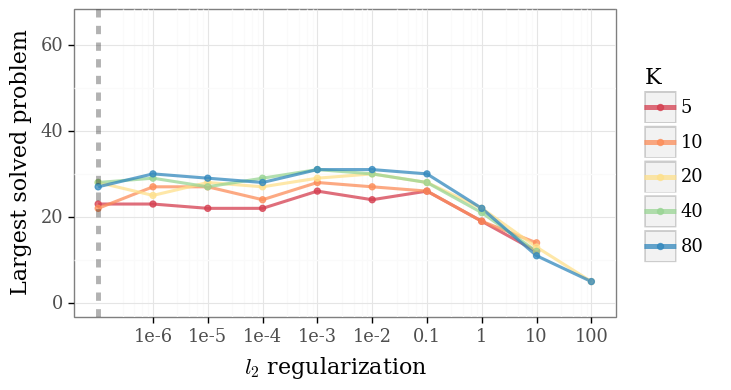}
    \vspace{-5mm}
    \caption{$l_2$ regularization has a very limited effect.}
    \label{fig:deep_sea_bsr_sweep}
\end{subfigure}
\quad
\centering
\begin{subfigure}{0.48\linewidth}
    \centering
    \includegraphics[width=\linewidth]{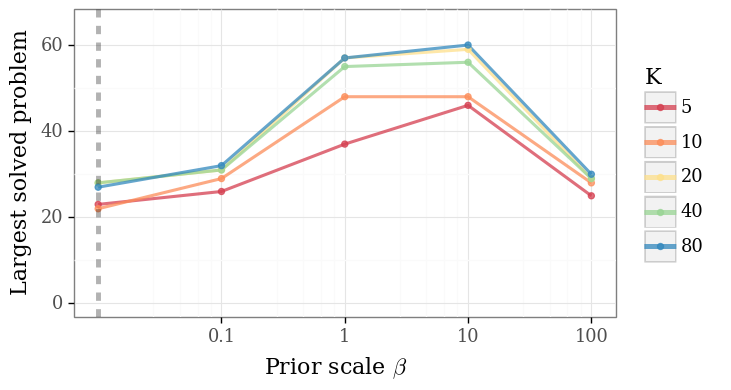}
    \vspace{-5mm}
    \caption{Additive prior greatly improves performance.}
    \label{fig:deep_sea_bsp_sweep}
\end{subfigure}
\caption{Comparing effects of different styles of prior regularization in Bootstrapped DQN.}
\label{fig:prior-1}
\end{figure}

Figure \ref{fig:prior-1} presents a more detailed analysis of the sensitivity of our approach to the tuning parameters of different regularization approaches.
We repeat the experiments of Figure \ref{fig:deep_sea_scale} and examine the size of the largest problem solved before $50$K episodes.
In each case larger ensembles lead to better performance, but this effect tends to plateau relatively early.
Figure \ref{fig:deep_sea_bsr_sweep} shows that regularization provides little or no benefit to BSR.
Figure \ref{fig:deep_sea_bsp_sweep} examines the effect of scaling the randomly initialized MLP by a scalar hyperparameter $\beta$.

\vspace{-4mm}
\subsubsubsection{\bf How does BSP solve this exponentially-difficult problem?}
\vspace{-1mm}

At first glance this `chain' problem may seem like an impossible task.
Finding the single rewarding policy out of $2^N$ is not simply a needle-in-a-haystack, but more akin to looking for a piece of hay in a needle-stack!
Since every policy apart from the rewarding one is painful, it's very tempting for an agent to give up and receive reward zero.
We now provide some intuition for how BSP is able to consistently and scalably solve such a difficult task.

One way to interpret this result is through analysing BSP with linear function approximation via Lemma \ref{thm:correctness}.
As outlined in Section \ref{sec:how_does_it_solve}, BSP with linear function approximation satisfies a polynomial regret bound \cite{osband2017deep}.
Further, this empirical scaling matches that predicted by the regret bound tabular domain \cite{osband2016posterior} (see Figure \ref{fig:learn_log_scale}).
Here, the prior function plays a crucial role - it provides motivation for the agent to explore even when the observed data has low (or no) reward.
Note that it is not necessary the sampled prior function leads to a good policy itself; in fact this is exponentially unlikely according to our initialization scheme.
The crucial element is that when a new state $s'$ is reached there is \textit{some} ensemble member that estimates $\max_{a'}Q_k(s',a')$ is sufficiently positive to warrant visiting, even if it causes some negative reward along the way.
In that case, when network $k$ is active it will seek out the potentially-informative $s'$ even if it is multiple timesteps away; this effect is sometimes called \textit{deep exploration}. 
We present an accompanying visualization at \url{http://bit.ly/rpf_nips}.

However, this connection to linear RLSVI does not inform why BSP should outperform BSR.
To account for this, we appeal to the functional dynamics of deep learning architectures (see Section \ref{sec:prior}).
In large networks weight decay (per BSR) may be an ineffective mechanism on the \textit{output} $Q$-values.
Instead, training an additive network via SGD (per BSP) may provide a more effective regularization on the output function \cite{zhang2016understanding,maennel2018gradient,bartlett2017spectrally}.
We expand on this hypothesis and further details of these experiments in Appendix \ref{app:deep_sea}.
This includes investigation of NoisyNets \cite{fortunato2017noisy} and dropout \cite{Gal2016Dropout}, which both perform poorly, and a comparison to UCB-based algorithms, which scale much worse than BSP, even with oracle access to state visit counts.

\vspace{-2mm}
\subsubsection{Cartpole swing-up}
\vspace{-1mm}
\label{sec:cartpole}

{\medmuskip=0mu
\thinmuskip=0mu
\thickmuskip=1mu
The experiments of Section \ref{sec:deep_sea} show that the choice of prior mechanism can be absolutely essential for efficient exploration via randomized value functions.
However, since the underlying system is a small finite MDP we might observe similar performance through a tabular algorithm.
In this section we investigate a classic benchmark problem that necessitates nonlinear function approximation: cartpole \cite{Sutton2017}.
We modify the classic formulation so that the pole begins hanging down and the agent only receives a reward when the pole is upright, balanced, and centered\footnote{We use the DeepMind control suite \cite{tassa2018deepmind} with reward $+1$ only when $\cos(\theta) > 0.95$, $|x| < 0.1$, $|\dot{\theta}| < 1$, and $|\dot{x}| < 1$. Each episode lasts 1,000 time steps, simulating 10 seconds of interaction.}.
We also add a cost of $0.1$ for moving the cart.
This problem embodies many of the same aspects of \ref{sec:deep_sea}, but since the agent interacts with the environment through state $s_t = (\cos(\theta_t), \sin(\theta_t), \dot{\theta_t}, x_t, \dot{x_t})$, the agent must also learn nonlinear generalization.
Tabular approaches are not practical due to the curse of dimensionality.
}

\begin{figure}[h!]
  \centering
  \begin{subfigure}{0.48\linewidth}
      \centering
      \includegraphics[width=\linewidth]{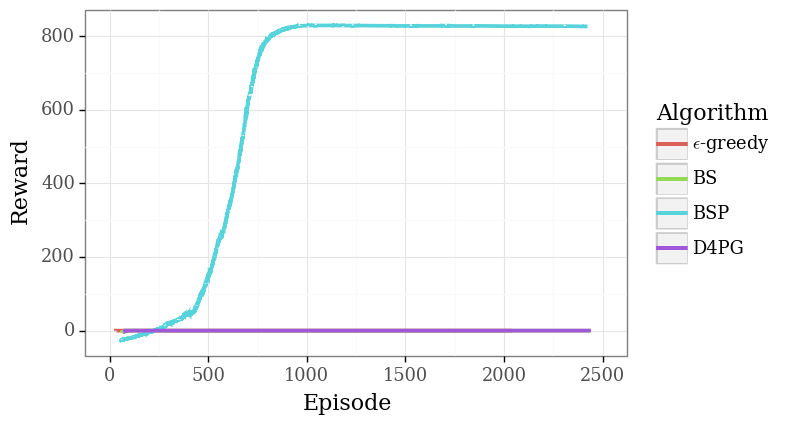}
      \vspace{-6mm}
      \caption{Only BSP learns a performant policy.}
      \label{fig:cartpole_full}
  \end{subfigure}
  \quad
  \begin{subfigure}{0.48\linewidth}
      \centering
      \includegraphics[width=\linewidth]{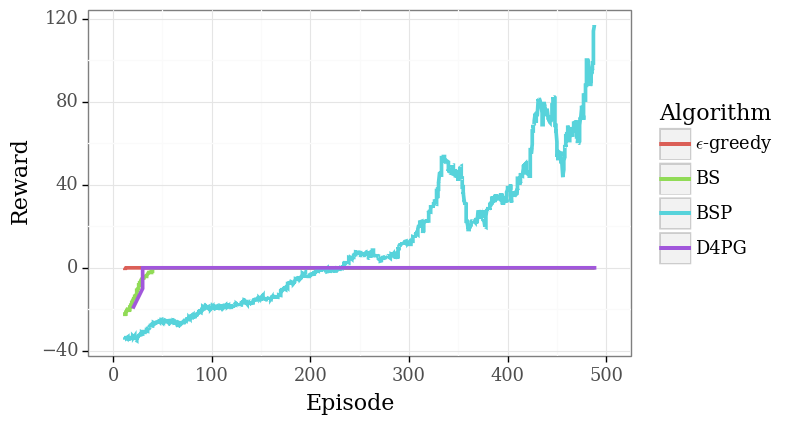}
      \vspace{-6mm}
      \caption{Inspecting the first 500 episodes.}
      \label{fig:cartpole_zoom}
  \end{subfigure}
  \caption{Learning curves for the modified cartpole swing-up task.}
  \label{fig:cartpole_plot}
\end{figure}
\vspace{-3mm}

Figure \ref{fig:cartpole_plot} compares the performance of DQN with $\epsilon$-greedy, bootstrap without prior (BS), bootstrap with prior networks (BSP) and the state-of-the-art continuous control algorithm D4PG, itself an application of `distributional RL' \cite{barth2018distributed}.
Only BSP learns a performant policy; no other approach ever attains any positive reward.
We push experimental details, including hyperparameter analysis, to Appendix \ref{app:cartpole}.
These results are significant in that they show that our intuitions translate from simple domains to more complex nonlinear settings, although the underlying state is relatively low dimensional.
Our next experiments investigate performance in a high dimensional and visually rich domain.

\subsubsection{Montezuma's revenge}
\label{sec:mz}

\begin{figure}[h!]
  \centering
  \includegraphics[width=\linewidth]{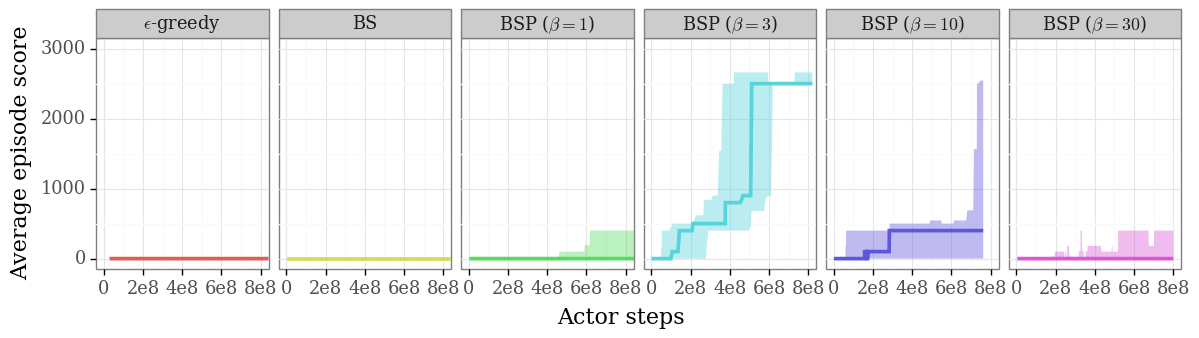}
  \vspace{-6mm}
  \caption{The prior network qualitatively changes behavior on Montezuma's revenge.}
  \label{fig:montezuma_plot}
\end{figure}

Our final experiment comes from the Arcade Learning Environment and the canonical sparse reward game, Montezuma's Revenge \cite{bellemare2013arcade}.
The agent interacts directly with the pixel values and, even under an optimal policy, there can be hundreds of time steps between rewarding actions.
This problem presents a significant exploration challenge in a visually rich environment; many published algorithms are essentially unable to attain any reward here \cite{mnih2015human,mnih2016asynchronous}.
We compare performance against a baseline distributed DQN agent with double Q-learning, prioritized experience replay and dueling networks \cite{horgan2018apex,hasselt2016doubledqn,schaul2015prioritized,wang2015duelling}.
To save computation we follow previous work and use a shared convnet for the ensemble uncertainty \cite{osband2016deep,azizzadenesheli2018efficient}.
Figure \ref{fig:montezuma_plot} presents the results for varying prior scale $\beta$ averaged over three seeds.
Once again, we see that the prior network can be absolutely critical to successful exploration.

\vspace{-2mm}
\section{Conclusion}
\vspace{-2mm}
\label{sec:conclusion}

This paper highlights the importance of uncertainty estimates in deep RL, the need for an effective `prior' mechanism, and its potential benefits towards efficient exploration.
We present some alarming shortcomings of existing methods and suggest bootstrapped ensembles with randomized prior functions as a simple, practical alternative.
We support our claims through an analysis of this method in the linear setting, together with a series of simple experiments designed to highlight the key issues.
Our work leaves several open questions.
What kinds of prior functions are appropriate for deep RL?
Can they be optimized or `meta-learned'?
Can we distill the ensemble process to a single network?
We hope this work helps to inspire solutions to these problems, and also build connections between the theory of efficient learning and practical algorithms for deep reinforcement learning.

\newpage
\textbf{Acknowledgements} \\
We would like to thank many people who made important contributions to this paper.
This paper can be thought of as a specific type of `deep exploration via randomized value functions', whose line of research has been crucially driven by the contributions of (and conversations with) Benjamin Van Roy, Daniel Russo and Zheng Wen.
Further, we would like to acknowledge the many helpful comments and support from Mohammad Gheshlaghi Azar, David Budden, David Silver and Justin Sirignano.
Finally, we would like to make a special mention for Hado Van Hasselt, who coined the term `hay in a needle-stack' to describe our experiments from Section \ref{sec:deep_sea}.

{
\small
\bibliographystyle{plain}
\bibliography{reference}

\begin{thebibliography}{10}

\bibitem{agrawal2012analysis}
Shipra Agrawal and Navin Goyal.
\newblock Analysis of {Thompson} sampling for the multi-armed bandit problem.
\newblock In {\em Conference on Learning Theory}, pages 39--1, 2012.

\bibitem{agrawal2012further}
Shipra Agrawal and Navin Goyal.
\newblock Further optimal regret bounds for {Thompson} sampling.
\newblock In {\em Artificial Intelligence and Statistics}, pages 99--107, 2013.

\bibitem{azizzadenesheli2018efficient}
Kamyar Azizzadenesheli, Emma Brunskill, and Animashree Anandkumar.
\newblock Efficient exploration through bayesian deep q-networks.
\newblock {\em arXiv preprint arXiv:1802.04412}, 2018.

\bibitem{barth2018distributed}
Gabriel Barth-Maron, Matthew~W Hoffman, David Budden, Will Dabney, Dan Horgan,
  Alistair Muldal, Nicolas Heess, and Timothy Lillicrap.
\newblock Distributed distributional deterministic policy gradients.
\newblock {\em arXiv preprint arXiv:1804.08617}, 2018.

\bibitem{bartlett2017spectrally}
Peter~L Bartlett, Dylan~J Foster, and Matus~J Telgarsky.
\newblock Spectrally-normalized margin bounds for neural networks.
\newblock In {\em Advances in Neural Information Processing Systems 30}, pages
  6241--6250, 2017.

\bibitem{bellemare2016countbased}
Marc Bellemare, Sriram Srinivasan, Georg Ostrovski, Tom Schaul, David Saxton,
  and Remi Munos.
\newblock Unifying count-based exploration and intrinsic motivation.
\newblock In {\em Advances in Neural Information Processing Systems 29}, pages
  1471--1479. 2016.

\bibitem{c51}
Marc~G Bellemare, Will Dabney, and R{\'e}mi Munos.
\newblock {A Distributional Perspective on Reinforcement Learning}.
\newblock {\em Proceedings of the 34th International Conference on Machine
  Learning (ICML)}, 2017.

\bibitem{bellemare2017distributional}
Marc~G Bellemare, Will Dabney, and R{\'e}mi Munos.
\newblock A distributional perspective on reinforcement learning.
\newblock In {\em International Conference on Machine Learning}, pages
  449--458, 2017.

\bibitem{bellemare2013arcade}
Marc~G Bellemare, Yavar Naddaf, Joel Veness, and Michael Bowling.
\newblock The arcade learning environment: An evaluation platform for general
  agents.
\newblock {\em J. Artif. Intell. Res.(JAIR)}, 47:253--279, 2013.

\bibitem{Bertsekas1996}
Dimitri~P. Bertsekas and John Tsitsiklis.
\newblock {\em Neuro-Dynamic Programming}.
\newblock Athena Scientific, September 1996.

\bibitem{blundell2015weight}
Charles Blundell, Julien Cornebise, Koray Kavukcuoglu, and Daan Wierstra.
\newblock Weight uncertainty in neural networks.
\newblock {\em arXiv preprint arXiv:1505.05424}, 2015.

\bibitem{cox1979theoretical}
David~Roxbee Cox and David~Victor Hinkley.
\newblock {\em Theoretical statistics}.
\newblock CRC Press, 1979.

\bibitem{dabney2017qr}
Will Dabney, Mark Rowland, Marc~G Bellemare, and R{\'e}mi Munos.
\newblock Distributional reinforcement learning with quantile regression.
\newblock In {\em Proceedings of the AAAI Conference on Artificial
  Intelligence}, 2018.

\bibitem{de1937prevision}
Bruno De~Finetti.
\newblock La pr{\'e}vision: ses lois logiques, ses sources subjectives.
\newblock In {\em Annales de l'institut Henri Poincar{\'e}}, volume~7, pages
  1--68, 1937.

\bibitem{efron1982jackknife}
Bradley Efron.
\newblock {\em The jackknife, the bootstrap and other resampling plans},
  volume~38.
\newblock SIAM, 1982.

\bibitem{efron1994introduction}
Bradley Efron and Robert~J Tibshirani.
\newblock {\em An introduction to the bootstrap}.
\newblock CRC press, 1994.

\bibitem{fortunato2017noisy}
Meire Fortunato, Mohammad~Gheshlaghi Azar, Bilal Piot, Jacob Menick, Ian
  Osband, Alex Graves, Vlad Mnih, Remi Munos, Demis Hassabis, Olivier Pietquin,
  et~al.
\newblock Noisy networks for exploration.
\newblock In {\em Proc. of ICLR}, 2018.

\bibitem{fushiki2005bootstrap}
Tadayoshi Fushiki.
\newblock Bootstrap prediction and bayesian prediction under misspecified
  models.
\newblock {\em Bernoulli}, pages 747--758, 2005.

\bibitem{fushiki2005nonparametric}
Tadayoshi Fushiki, Fumiyasu Komaki, Kazuyuki Aihara, et~al.
\newblock Nonparametric bootstrap prediction.
\newblock {\em Bernoulli}, 11(2):293--307, 2005.

\bibitem{Gal2016Dropout}
Yarin Gal and Zoubin Ghahramani.
\newblock Dropout as a {B}ayesian approximation: Representing model uncertainty
  in deep learning.
\newblock In {\em International Conference on Machine Learning}, 2016.

\bibitem{gal2017concrete}
Yarin Gal, Jiri Hron, and Alex Kendall.
\newblock Concrete dropout.
\newblock In {\em Advances in Neural Information Processing Systems}, pages
  3584--3593, 2017.

\bibitem{gal2016improving}
Yarin Gal, Rowan McAllister, and Carl~Edward Rasmussen.
\newblock Improving pilco with bayesian neural network dynamics models.
\newblock In {\em Data-Efficient Machine Learning workshop, ICML}, 2016.

\bibitem{glorot2010understanding}
Xavier Glorot and Yoshua Bengio.
\newblock Understanding the difficulty of training deep feedforward neural
  networks.
\newblock In {\em Proceedings of the 13th international conference on
  artificial intelligence and statistics}, pages 249--256, 2010.

\bibitem{hasselt2016doubledqn}
Hado~van Hasselt, Arthur Guez, and David Silver.
\newblock Deep reinforcement learning with double q-learning.
\newblock In {\em Proceedings of the Thirtieth AAAI Conference on Artificial
  Intelligence}, AAAI'16, pages 2094--2100. AAAI Press, 2016.

\bibitem{horgan2018apex}
Daniel Horgan, John Quan, David Budden, Gabriel Barth-Maron, Matteo Hessel,
  Hado~Van Hasselt, and David Silver.
\newblock Distributed prioritized experience replay.
\newblock In {\em 6th International Conference on Learning Represenations},
  2018.

\bibitem{jaksch2010near}
Thomas Jaksch, Ronald Ortner, and Peter Auer.
\newblock Near-optimal regret bounds for reinforcement learning.
\newblock {\em Journal of Machine Learning Research}, 11(Apr):1563--1600, 2010.

\bibitem{kearns2002near}
M.~Kearns and S.~Singh.
\newblock Near-optimal reinforcement learning in polynomial time.
\newblock {\em Machine Learning}, 49, 2002.

\bibitem{kingma2014adam}
Diederik Kingma and Jimmy Ba.
\newblock {Adam: A Method for Stochastic Optimization}.
\newblock {\em Proceedings of the International Conference on Learning
  Representations}, 2015.

\bibitem{kingma2013auto}
Diederik~P Kingma and Max Welling.
\newblock Auto-encoding variational bayes.
\newblock {\em International Conference on Learning Representations}, 2014.

\bibitem{krizhevsky2012imagenet}
Alex Krizhevsky, Ilya Sutskever, and Geoffrey~E Hinton.
\newblock Imagenet classification with deep convolutional neural networks.
\newblock In {\em Advances in Neural Information Processing Systems 25}, pages
  1097--1105, 2012.

\bibitem{lakshminarayanan2017simple}
Balaji Lakshminarayanan, Alexander Pritzel, and Charles Blundell.
\newblock Simple and scalable predictive uncertainty estimation using deep
  ensembles.
\newblock In {\em Advances in Neural Information Processing Systems}, pages
  6405--6416, 2017.

\bibitem{lecun2015deep}
Yann LeCun, Yoshua Bengio, and Geoffrey Hinton.
\newblock Deep learning.
\newblock {\em Nature}, 521(7553):436, 2015.

\bibitem{legg2007collection}
Shane Legg, Marcus Hutter, et~al.
\newblock A collection of definitions of intelligence.
\newblock {\em Frontiers in Artificial Intelligence and applications}, 157:17,
  2007.

\bibitem{leike2016thompson}
Jan Leike, Tor Lattimore, Laurent Orseau, and Marcus Hutter.
\newblock Thompson sampling is asymptotically optimal in general environments.
\newblock {\em Uncertainty in Artificial Intelligence}, 2016.

\bibitem{lipton2016efficient}
Zachary~C Lipton, Jianfeng Gao, Lihong Li, Xiujun Li, Faisal Ahmed, and
  Li~Deng.
\newblock Efficient exploration for dialogue policy learning with bbq networks
  \& replay buffer spiking.
\newblock {\em arXiv preprint arXiv:1608.05081}, 2016.

\bibitem{lu2017ensemble}
Xiuyuan Lu and Benjamin Van~Roy.
\newblock Ensemble sampling.
\newblock In {\em Advances in Neural Information Processing Systems}, pages
  3260--3268, 2017.

\bibitem{mackay1992practical}
David~JC MacKay.
\newblock A practical {B}ayesian framework for backpropagation networks.
\newblock {\em Neural computation}, 4(3):448--472, 1992.

\bibitem{maennel2018gradient}
Hartmut Maennel, Olivier Bousquet, and Sylvain Gelly.
\newblock Gradient descent quantizes {R}e{LU} network features.
\newblock {\em arXiv preprint arXiv:1803.08367}, 2018.

\bibitem{mania2018simple}
Horia Mania, Aurelia Guy, and Benjamin Recht.
\newblock Simple random search provides a competitive approach to reinforcement
  learning.
\newblock {\em arXiv preprint arXiv:1803.07055}, 2018.

\bibitem{mihatsch2002risk}
Oliver Mihatsch and Ralph Neuneier.
\newblock Risk-sensitive reinforcement learning.
\newblock {\em Machine learning}, 49(2-3):267--290, 2002.

\bibitem{mnih2016asynchronous}
Volodymyr Mnih, Adria~Puigdomenech Badia, Mehdi Mirza, Alex Graves, Timothy
  Lillicrap, Tim Harley, David Silver, and Koray Kavukcuoglu.
\newblock Asynchronous methods for deep reinforcement learning.
\newblock In {\em Proc. of ICML}, 2016.

\bibitem{mnih2015human}
Volodymyr Mnih, Koray Kavukcuoglu, David Silver, Andrei~A Rusu, Joel Veness,
  Marc~G Bellemare, Alex Graves, Martin Riedmiller, Andreas~K Fidjeland, Georg
  Ostrovski, et~al.
\newblock Human-level control through deep reinforcement learning.
\newblock {\em Nature}, 518(7540):529--533, 2015.

\bibitem{neal2012bayesian}
Radford~M Neal.
\newblock {\em Bayesian learning for neural networks}, volume 118.
\newblock Springer Science \& Business Media, 2012.

\bibitem{o2017uncertainty}
Brendan O'Donoghue, Ian Osband, Remi Munos, and Volodymyr Mnih.
\newblock The uncertainty bellman equation and exploration.
\newblock {\em arXiv preprint arXiv:1709.05380}, 2017.

\bibitem{osband2016}
Ian Osband.
\newblock {\em Deep Exploration via Randomized Value Functions}.
\newblock PhD thesis, Stanford University, 2016.

\bibitem{osband2016risk}
Ian Osband.
\newblock Risk versus uncertainty in deep learning: Bayes, bootstrap and the
  dangers of dropout.
\newblock 2016.

\bibitem{osband2016deep}
Ian Osband, Charles Blundell, Alexander Pritzel, and Benjamin Van~Roy.
\newblock Deep exploration via bootstrapped {DQN}.
\newblock In {\em Advances In Neural Information Processing Systems 29}, pages
  4026--4034, 2016.

\bibitem{Osband2013}
Ian Osband, Daniel Russo, and Benjamin Van~Roy.
\newblock {(More)} efficient reinforcement learning via posterior sampling.
\newblock In {\em Advances in Neural Information Processing Systems 26}, pages
  3003--3011. 2013.

\bibitem{osband2017deep}
Ian Osband, Daniel Russo, Zheng Wen, and Benjamin Van~Roy.
\newblock Deep exploration via randomized value functions.
\newblock {\em arXiv preprint arXiv:1703.07608}, 2017.

\bibitem{osband2015bootstrapped}
Ian Osband and Benjamin Van~Roy.
\newblock Bootstrapped {T}hompson sampling and deep exploration.
\newblock {\em arXiv preprint arXiv:1507.00300}, 2015.

\bibitem{osband2016posterior}
Ian Osband and Benjamin Van~Roy.
\newblock Why is posterior sampling better than optimism for reinforcement
  learning?
\newblock In {\em Proceedings of the 34th International Conference on Machine
  Learning}, pages 2701--2710, 2017.

\bibitem{osband2016rlsvi}
Ian Osband, Benjamin Van~Roy, and Zheng Wen.
\newblock Generalization and exploration via randomized value functions.
\newblock In {\em Proceedings of The 33rd International Conference on Machine
  Learning}, pages 2377--2386, 2016.

\bibitem{ostrovski2017countbased}
Georg Ostrovski, Marc~G Bellemare, Aaron van~den Oord, and R\'emi Munos.
\newblock Count-based exploration with neural density models.
\newblock In {\em Proc. of ICML}, 2017.

\bibitem{plappert2017parameter}
Matthias Plappert, Rein Houthooft, Prafulla Dhariwal, Szymon Sidor, Richard~Y
  Chen, Xi~Chen, Tamim Asfour, Pieter Abbeel, and Marcin Andrychowicz.
\newblock Parameter space noise for exploration.
\newblock {\em arXiv preprint arXiv:1706.01905}, 2017.

\bibitem{rumelhart1985learning}
David~E Rumelhart, Geoffrey~E Hinton, and Ronald~J Williams.
\newblock Learning internal representations by error propagation.
\newblock Technical report, DTIC Document, 1985.

\bibitem{RusmevichientongT2010}
Paat Rusmevichientong and John~N. Tsitsiklis.
\newblock Linearly parameterized bandits.
\newblock {\em Math. Oper. Res.}, 35(2):395--411, 2010.

\bibitem{Russo2014}
Daniel Russo and Benjamin Van~Roy.
\newblock Learning to optimize via posterior sampling.
\newblock {\em Mathematics of Operations Research}, 39(4):1221--1243, 2014.

\bibitem{russo2017tutorial}
Daniel Russo, Benjamin Van~Roy, Abbas Kazerouni, and Ian Osband.
\newblock A tutorial on {T}hompson sampling.
\newblock {\em arXiv preprint arXiv:1707.02038}, 2017.

\bibitem{schaul2015prioritized}
Tom Schaul, John Quan, Ioannis Antonoglou, and David Silver.
\newblock Prioritized experience replay.
\newblock {\em CoRR}, abs/1511.05952, 2015.

\bibitem{silver2016alphago}
David Silver, Aja Huang, Chris~J Maddison, Arthur Guez, Laurent Sifre, George
  Van Den~Driessche, Julian Schrittwieser, Ioannis Antonoglou, Veda
  Panneershelvam, Marc Lanctot, et~al.
\newblock Mastering the game of go with deep neural networks and tree search.
\newblock {\em Nature}, 529(7587):484--489, 2016.

\bibitem{srivastava2014dropout}
Nitish Srivastava, Geoffrey Hinton, Alex Krizhevsky, Ilya Sutskever, and Ruslan
  Salakhutdinov.
\newblock Dropout: A simple way to prevent neural networks from overfitting.
\newblock {\em The Journal of Machine Learning Research}, 15(1):1929--1958,
  2014.

\bibitem{strens2000bayesian}
Malcolm Strens.
\newblock A {B}ayesian framework for reinforcement learning.
\newblock In {\em International Conference on Machine Learning}, pages
  943--950, 2000.

\bibitem{Sutton2017}
Richard Sutton and Andrew Barto.
\newblock {\em Reinforcement Learning: An Introduction}.
\newblock MIT Press, 2017.

\bibitem{sutton2011horde}
Richard~S Sutton, Joseph Modayil, Michael Delp, Thomas Degris, Patrick~M
  Pilarski, Adam White, and Doina Precup.
\newblock Horde: A scalable real-time architecture for learning knowledge from
  unsupervised sensorimotor interaction.
\newblock In {\em The 10th International Conference on Autonomous Agents and
  Multiagent Systems-Volume 2}, pages 761--768. International Foundation for
  Autonomous Agents and Multiagent Systems, 2011.

\bibitem{tang2017variational}
Yunhao Tang and Alp Kucukelbir.
\newblock Variational deep q network.
\newblock {\em arXiv preprint arXiv:1711.11225}, 2017.

\bibitem{tassa2018deepmind}
Yuval Tassa, Yotam Doron, Alistair Muldal, Tom Erez, Yazhe Li, Diego de~Las
  Casas, David Budden, Abbas Abdolmaleki, Josh Merel, Andrew Lefrancq, et~al.
\newblock Deepmind control suite.
\newblock {\em arXiv preprint arXiv:1801.00690}, 2018.

\bibitem{tesauro1995temporal}
Gerald Tesauro.
\newblock Temporal difference learning and {TD}-gammon.
\newblock {\em Communications of the ACM}, 38(3):58--68, 1995.

\bibitem{Thompson1933}
William~R Thompson.
\newblock On the likelihood that one unknown probability exceeds another in
  view of the evidence of two samples.
\newblock {\em Biometrika}, 25(3/4):285--294, 1933.

\bibitem{touati2018randomized}
Ahmed Touati, Harsh Satija, Joshua Romoff, Joelle Pineau, and Pascal Vincent.
\newblock Randomized value functions via multiplicative normalizing flows.
\newblock {\em arXiv preprint arXiv:1806.02315}, 2018.

\bibitem{van2016wavenet}
Aaron Van Den~Oord, Sander Dieleman, Heiga Zen, Karen Simonyan, Oriol Vinyals,
  Alex Graves, Nal Kalchbrenner, Andrew Senior, and Koray Kavukcuoglu.
\newblock Wavenet: A generative model for raw audio.
\newblock {\em arXiv preprint arXiv:1609.03499}, 2016.

\bibitem{wald1992statistical}
Abraham Wald.
\newblock Statistical decision functions.
\newblock In {\em Breakthroughs in Statistics}, pages 342--357. Springer, 1992.

\bibitem{wang2015duelling}
Ziyu Wang, Nando de~Freitas, and Marc Lanctot.
\newblock Dueling network architectures for deep reinforcement learning.
\newblock {\em CoRR}, abs/1511.06581, 2015.

\bibitem{zhang2016understanding}
Chiyuan Zhang, Samy Bengio, Moritz Hardt, Benjamin Recht, and Oriol Vinyals.
\newblock Understanding deep learning requires rethinking generalization.
\newblock {\em CoRR}, abs/1611.03530, 2016.

\end{thebibliography}
}

\newpage
\appendix

\section{Reinforcement learning algorithm}
\label{app:algorithm}

In this appendix we fill out the details for the complete pseudocode for the BootDQN+Prior RL agent.
Our problem setting matches the description of prior work \cite{osband2017deep}; we reproduce the algorithms and figures in this section for the convenience of our readers.
At a high level, our agents interact with the environment through repeated finite episodes as described by Algorithm \ref{alg:live}.
To describe an agent, we must simply implement the \texttt{act}, \texttt{update\_buffer} and \texttt{learn\_from\_buffer} methods.

\begin{algorithm}[!htpb]
\caption{$\mathtt{live}$}
\label{alg:live}
\begin{tabular}{lll}
\textbf{Input:} & \texttt{agent} & methods $\mathtt{act}, \mathtt{update\_buffer}, \mathtt{learn\_from\_buffer}$ \\
                & \texttt{environment} & methods $\mathtt{reset}, \mathtt{step}$ \\
\end{tabular}
\begin{algorithmic}[1]
\For{episode $= 1,2,\ldots$}
\State \texttt{agent.learn\_from\_buffer()}
\State \texttt{transition} $\leftarrow$ \texttt{environment.reset()}
\While{\texttt{transition.new\_state} is not {\bf null}}
\State \texttt{action} $\leftarrow$ \texttt{agent.act(transition.new\_state)}
\State \texttt{transition} $\leftarrow$ \texttt{environment.step(action)}
\State \texttt{agent.update\_buffer(transition)}
\EndWhile
\EndFor
\end{algorithmic}
\end{algorithm}

\begin{figure}[!h]
\centering
\begin{subfigure}{.5\textwidth}
  \centering
  \includegraphics[width=.95\linewidth]{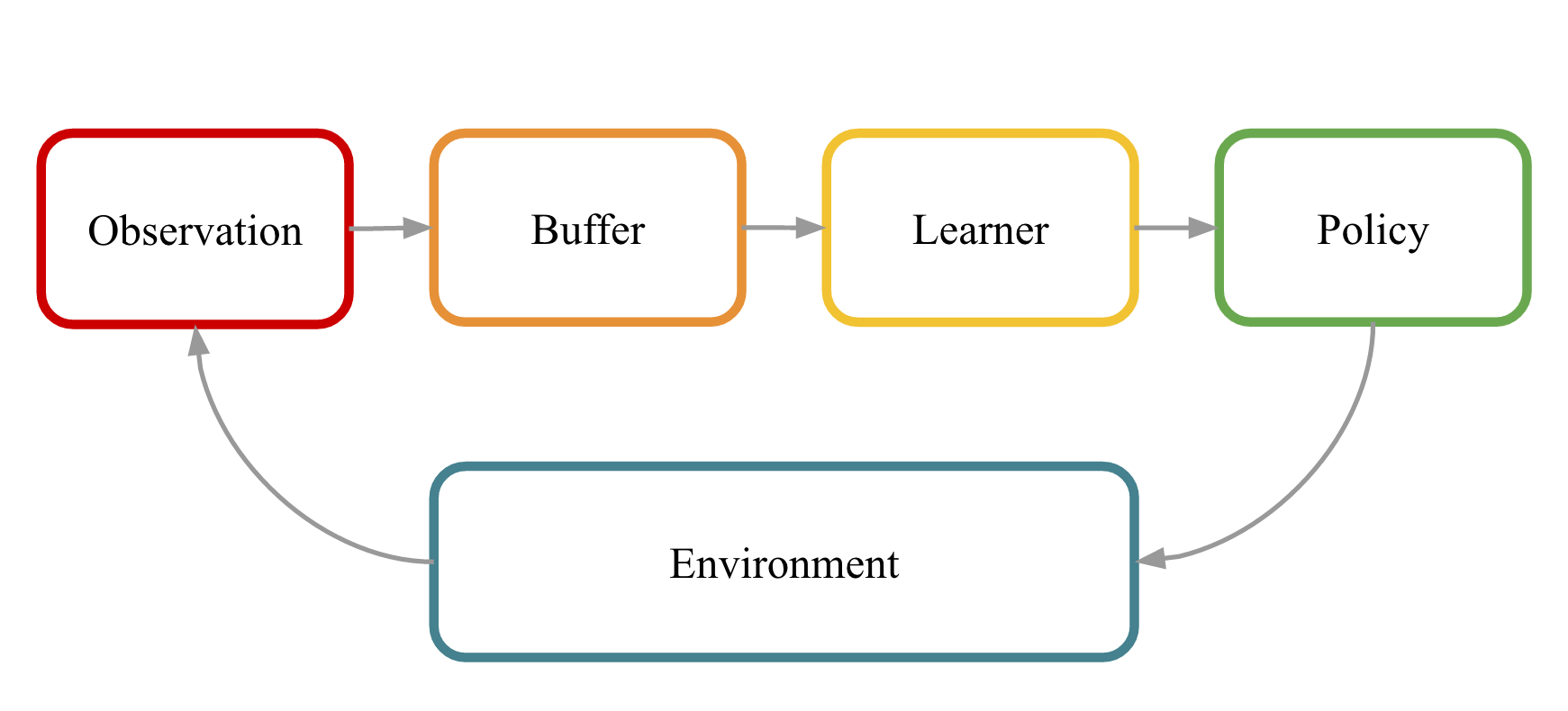}
  \vspace{-2mm}
  \caption{Learning a single value function}
  \label{fig:online_lsvi}
\end{subfigure}%
\begin{subfigure}{.5\textwidth}
  \centering
  \includegraphics[width=.95\linewidth]{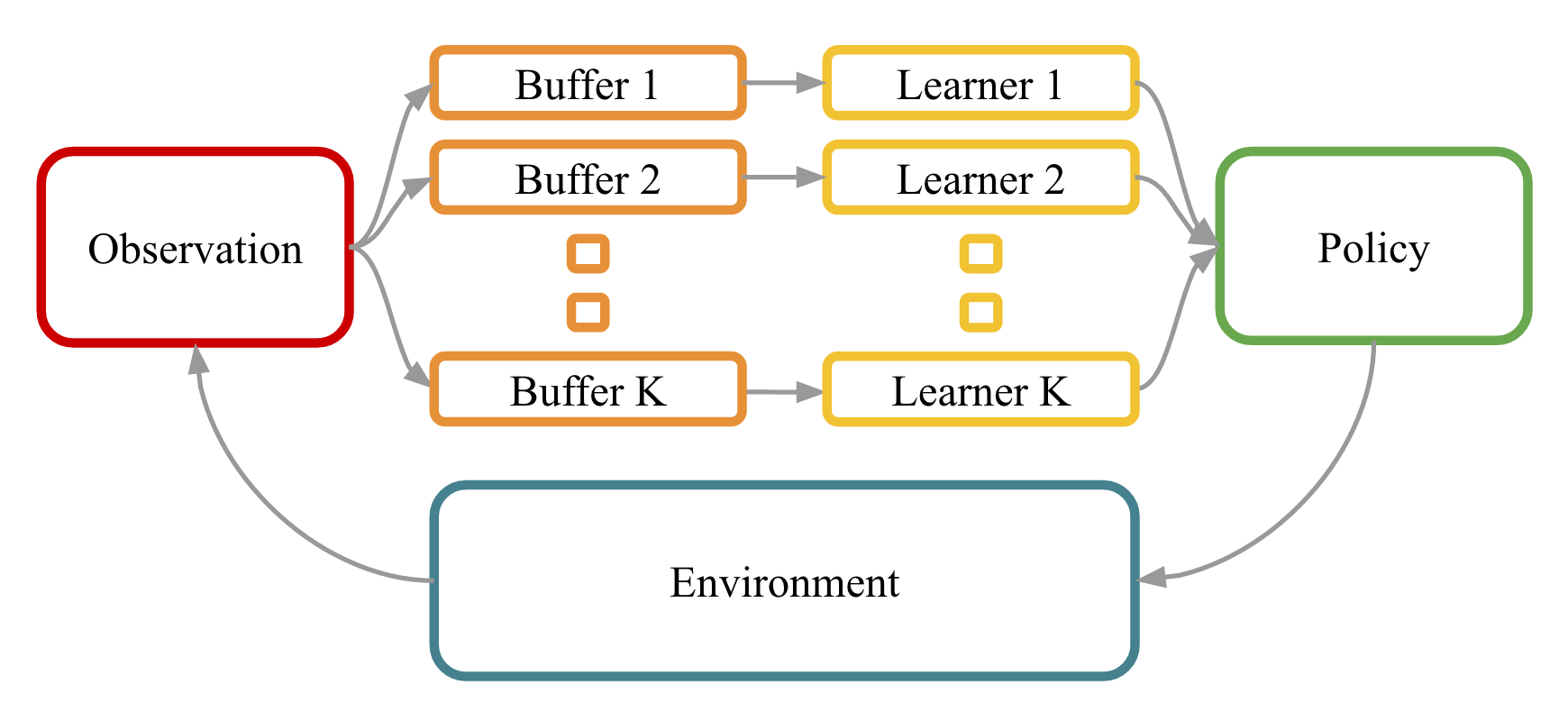}
  \vspace{-2mm}
  \caption{Learning multiple value functions in parallel}
  \label{fig:online_rlsvi}
\end{subfigure}
\caption{\small RLSVI via ensemble sampling, each member produced by LSVI on perturbed data.}
\label{fig:online_models}
\end{figure}

BootDQN+prior implements an $\mathtt{ensemble\_buffer}$ that maintains $K$ buffers in parallel, although this may clearly be implemented in an efficient way that uses $o(K)$ memory.
Figure \ref{fig:online_models} provides an illustration of how BootDQN learns and maintains $K$ estimates of the value function in parallel.

To implement an online double-or-nothing bootstrap we employ Algorithm \ref{alg:ensemble_update_bootstrap}, which assigns each transition to each ensemble buffer with probability $\frac{1}{2}$.

\begin{algorithm}[!htpb]
\caption{$\mathtt{ensemble\_buffer.update\_bootstrap}(\cdot)$}
\label{alg:ensemble_update_bootstrap}

\begin{tabular}{lll}
\textbf{Input:} & $\mathtt{transition}$ & $(s_t, a_t, r_t, s'_t, t)$ \\
\textbf{Updates:} & $\mathtt{ensemble\_buffer}$ & replay buffer of $K$-parallel perturbed data
\end{tabular}

\begin{algorithmic}[1]
\For{$k$ in $(1,\ldots,K)$}
\If{$m^k_t \sim {\rm Unif}(\{0, 1\}) = 1$}
\State $\mathtt{ensemble\_buffer[k].enqueue(}(s_t, a_t,  r_t, s'_t, t))$
\EndIf
\EndFor
\end{algorithmic}
\end{algorithm}

Algorithm \ref{alg:learn_ensemble_rlsvi} describes the $\mathtt{learn\_from\_buffer}$ method for the agent.
For our experiments, we sometimes amend Algorithm \ref{alg:live} to learn periodically every $N$ steps, rather than only at the end of the episode, but we mention this in the text where this is the case.
This practice is common for most implementations of DQN and other reinforcement learning algorithms, but it does not play a significant role in our algorithm.

The final piece to describe BootDQN+prior is the $\mathtt{act}$ method for action selection.
We employ a form of approximate Thompson sampling for RL via randomized value functions.
Every episode, the agent selects $j \sim {\rm Unif}(1,..,K)$ and follows the greedy policy for $Q_j$ for the duration of the episode.

\section{Reinforcement learning experiments}
\label{app:experiment_details}

In this section, we expand on details for the experimental set-up together with some additional results.
Unless otherwise stated we use TensorFlow defaults, Adam optimizer with learning rate $10^{-3}$ and uniform experience replay with batch size $128$.
For our $\epsilon$-greedy DQN baseline, we anneal epsilon linearly over $2000$ episodes and perform hyperparameter sweeps over the initial epsilon $\epsilon_0$.
All other agents (NoisyNet, Dropout, Ensemble, Bootstrap) use greedy policies according to an appropriate per-episode Thompson sampling.


\subsection{Chain environments}
\label{app:deep_sea}

\begin{figure}[h!]
  \centering
  \includegraphics[width=\linewidth]{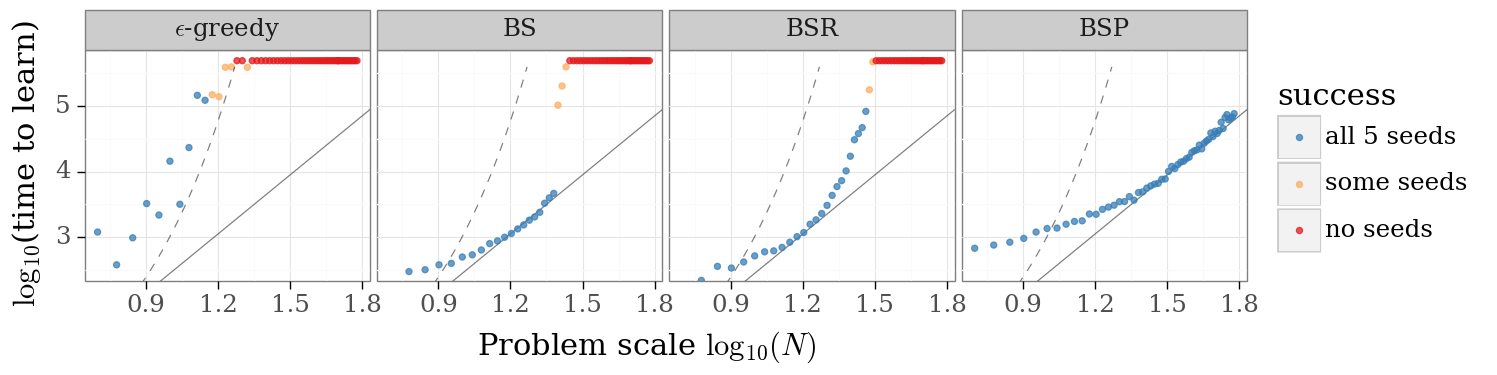}
  \vspace{-2mm}
  \caption{Log-log plot demonstrates scaling of learning behaviour.}
  \label{fig:learn_log_scale}
\end{figure}

Figure \ref{fig:deep_sea_scale} shows the time it takes each agent to learn a problem of size $ N $.
Figure \ref{fig:learn_log_scale} reproduces these results but on a log-log scale, which helps to reveal the problem scaling as $N$ increases.
As in Figure \ref{fig:deep_sea_scale}, the dashed line corresponds to a dithering lower bound $T_{\rm learn} = 2^N$.
We also include a solid line with slope equal to three, corresponding to a polynomial growth $T_{\rm learn} = \tilde{O}(N^3)$.

\vspace{-2mm}

\begin{figure}[H]
  \centering
  \includegraphics[width=0.7\linewidth]{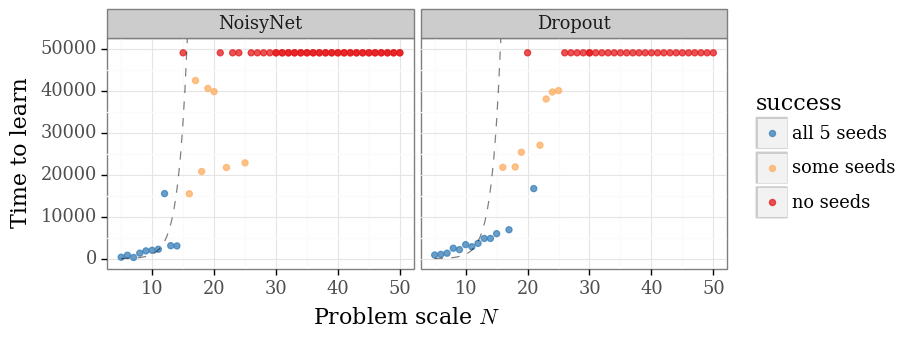}
  \vspace{-2mm}
  \caption{Learning time for noisy and dropout; neither approach scales well.}
  \label{fig:deep_sea_compare_noisy_dropout}
\end{figure}

In addition to BSP, BSR, BS and $ \epsilon $-greedy displayed in Figure \ref{fig:deep_sea_scale}, we also ran parameter sweeps for dropout, NoisyNet and a count-based exploration strategy.
Figure \ref{fig:deep_sea_compare_noisy_dropout} presents the result for NoisyNet and dropout, each individually tuned up to 50k episodes.
Even after tuning dropout rate and sampling frequency (by episode or by timestep) neither dropout nor NoisyNet scale successfuly to large domains.

\vspace{-2mm}

\begin{figure}[H]
  \centering
  \includegraphics[width=0.99\linewidth]{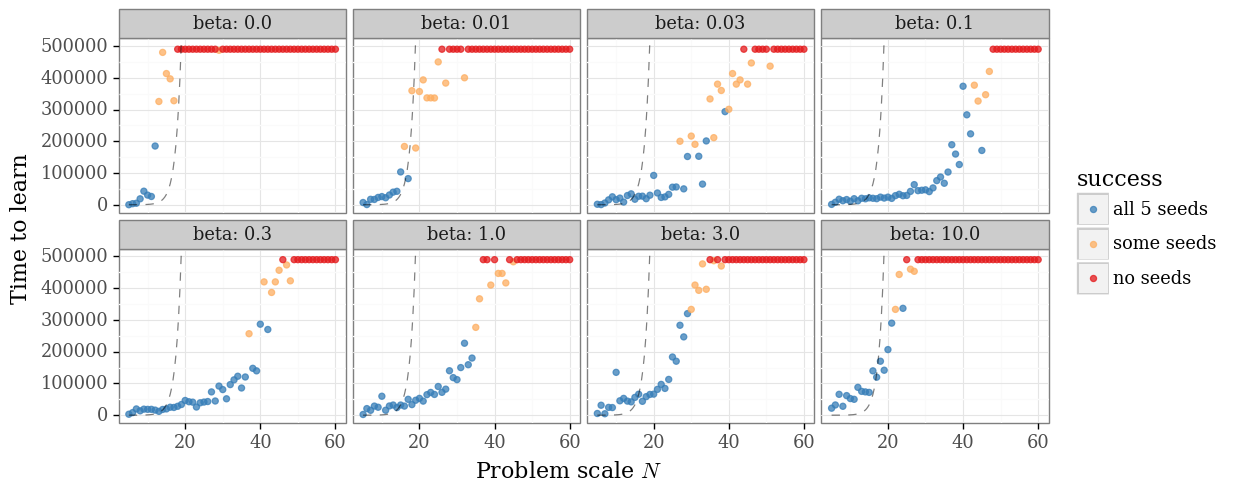}
  \vspace{-2mm}
  \caption{Sweeping over optimistic bonus; no scale of $\beta$ matches BSP performance.}
  \label{fig:deep_sea_compare_ucb}
\end{figure}

To compare with `count-based' exploration we implement a version of DQN that optimizes the true reward plus a UCB exploration bonus $\frac{\beta}{\sqrt{N_t(s)}}$, where $N(s)$ is the number of visits to state $s$ prior to time $t$ \cite{jaksch2010near,bellemare2016countbased}.
Figure \ref{fig:deep_sea_compare_ucb} shows that this count-based exploration strategy performs much worse than BSP, even after sweeping over bonus scale $\beta$ and even with access to the true state visit-counts.
This mirrors the outperformance of PSRL vs UCRL in tabular reinfocement learning.
One explanation for this discrepancy comes from the inefficient way UCB-style algorithms propagate uncertainty over many timesteps \cite{osband2016posterior,o2017uncertainty}.

\vspace{-3mm}

For all of our algorithms we tune agent hyperparameters by grid search.
These were:
\begin{itemize}[noitemsep, nolistsep]
  \item {\textbf{$\mathbf{\epsilon}$-greedy}: $\epsilon = 0.1$, linearly annealed to zero.}
  \item {\textbf{BSP}: prior scale $\beta=10$ (Figure \ref{fig:deep_sea_bsp_sweep})}.
  \item {\textbf{BSR}: $l_2$ regularizer scale $\lambda = 0.1$ (Figure \ref{fig:deep_sea_bsr_sweep})}.
  \item {\textbf{Dropout}: Resample mask every step with $p_{\text{keep}} = 0.1$}.
  \item {\textbf{NoisyNet}: Resample noise every step.}
  \item {\textbf{UCB}: Optimistic bonus $\beta=0.1$ (Figure \ref{fig:deep_sea_compare_ucb}).}
\end{itemize}

\subsection{Sparse cartpole swing-up}
\label{app:cartpole}

\begin{figure}[h!]
  \centering
  \includegraphics[width=.6\linewidth]{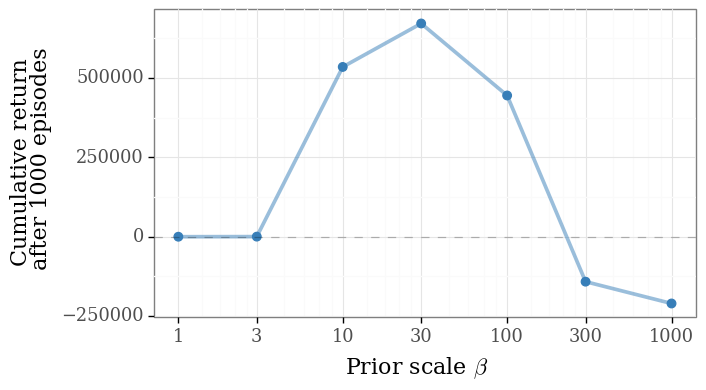}
  \caption{Sensitivity of performance to prior scale $\beta$.}
  \label{fig:cartpole_beta_sweep}
\end{figure}

In Section \ref{sec:cartpole} we presented experiments showing that BSP outperforms benchmark algorithms.
Figure \ref{fig:cartpole_beta_sweep} presents the sensitivity of BSP sensitivity to the prior scale $\beta$ on this domain.
Small values of $\beta$ prematurely and suboptimally converge to the stationary policy, and so receive zero cumulative reward.
Larger values of $\beta$ take longer to wash away their prior effect, but we expect them to learn a performant policy eventually.
This behaviour mirrors the scaling we saw in the chain environments, which is reassuring.

\subsection{Montezuma's revenge}

In our experiments, we use the standard Atari configuration and preprocessing including greyscaling, frame stacking, action repeats, and random no-op starts \cite{mnih2015human}, and the same agent hyperparameters as those used in the Ape-X paper \cite{horgan2018apex}.
However, our agent implementation is somewhat different and so our baseline results are not directly comparable across all games.

\section{Why do we need a `prior' mechanism for deep RL?}
\label{app:existing_bad}

Section \ref{sec:why_prior} outlines the need for a prior mechanism in deep RL, together with key failure cases for several of the most popular approaches.
Due to space limitations we provide only simple illustrations of potential inadequacies of each method and this does not preclude their efficacy on any particular domain.
In this appendix we expand on the details provided in Section \ref{sec:why_prior} and provide suggestions for how these approaches might be remedied by future work.

\subsection{Dropout as posterior approximation}
\label{app:dropout}

Previous work has suggested that dropout works as an effective variational approximation to the Bayesian posterior in neural networks, without special consideration for the network architecture \cite{Gal2016Dropout}.
However, Lemma \ref{thm:dropout} is a general statement that gives us cause to question the quality of this approximation.
In this subsection we dig deeper into an extremely simple estimation problem, a linear network with $d$ units to estimate the mean of a random variable $Y \in \Real$.
Even in this simple setting dropout performs poorly as a Bayesian posterior.

We form predictions $f_\theta = \sum_{i=1}^d w_i \theta_i$ with $w_i \sim {\rm Ber}(p)$, square loss and regularizer $\Rc(\theta) = \lambda \|\theta\|^2$ for $\lambda > 0$.
Then for any data $\Dc$ with empirical mean $\overline{y}$,  the expected loss solution to \eqref{eq:dropout} is given by \cite{srivastava2014dropout}
$\theta^*_p = \overline{\theta} \Ind \text{ for } \overline{\theta} = \frac{\overline{y}}{1 + p(d-1) + \frac{\lambda}{d}}.$ \footnote{This corrects an errant derivation in \cite{osband2016risk}, but maintains the same overall message.}
{\medmuskip=0mu
\thinmuskip=0mu
\thickmuskip=1mu
The resultant predictive distribution therefore has mean $\mu = \overline{\theta}dp$ and standard deviation $\sigma = \overline{\theta} \sqrt{d p (1-p)}$.

If we are to understand dropout as an approximation to a Bayesian posterior, then we should note that this behavior is unusual.
First, the only connection to the data is through the empirical mean $\overline{y}$; any possible dataset with the same mean would result in the same `posterior' distribution.
Second we note that $\sigma = \mu \sqrt{(1-p)/dp}$.
This coupling means it is not possible for $\sigma \rightarrow 0$ and $\mu \nrightarrow 0$, regardless of $\lambda$.
More typically we would imagine $\mu \rightarrow \Exp[Y]$ and $\sigma \rightarrow 0$ according to the Bayesian central limit theorem \cite{cox1979theoretical}.

This disconnect is not simply an analytical mistake, but can lead to arbitrarily bad decisions in even the simplest problems.
Imagine a simple two-armed bandit problem with one arm's rewards $\sim {\rm Ber}(1/2)$ and the other's $\sim {\rm Ber}(1/2 + \epsilon)$, and the agent does not know which arm is which a priori.
This style of problem is particularly well understood with guarantees that Thompson sampling with more reasonable forms of posterior approximation incur regret $\tilde{O}(\log(T))$ in this setting \cite{agrawal2012further}.
We refer to this problem as $\dagger$.
The following result highlights that dropout as posterior approximation can perform poorly even on this simple domain.

\begin{lemma}[Dropout sampling attains linear regret in $\dagger$]
\label{lem:dropout_explore}
\hspace{0.000001mm} \newline
Fix any $d$, $p$, $\lambda$ and consider the problem of $\dagger$ with an agent employing Thompson sampling by dropout for action selection.
Then the expected regret is $\Omega(T).$
\begin{proof}
For any $d$, $p$, $\lambda$ and any observed data $\Hc_t$, there exists a non-zero probability $P_1(s, p, \lambda, \Hc_t) > \frac{p^{d}}{2}$ of selecting action $1$ over action action $2$.
We can see this by imagining all units estimating action $2$ are set to zero, then there is at least 50\% chance of selection action $1$.
This proves\footnote{Note that this lower bound is very conservative and provided only for illustration. A more precise analysis would show poor performance even for large $d$.} that $ \Exp\left[ {\rm Regret}(T) \right] \ge \frac{\epsilon p^d}{2} T$ for all $T$.
\end{proof}
\end{lemma}

Although our analysis of dropout has focused on an exceedingly simple functional form, the key insight that the degree of variability in the posterior distribution does not concentrate with data extends to any neural network architecture.
Figure \ref{fig:dropout_no_converge} presents the dropout posterior on a simple regression task with a (20,20)-MLP with rectified linear units.
We display the predictive distribution under varying amounts of data.
The dropout sampling distribution does not converge with increasing amounts of data, whereas the bootstrapped sampling approach behaves much more reasonably.
This leads to poor performance in reinforcement learning tasks too, as we saw in Appendix \ref{app:experiment_details}.

\begin{figure}[h!]
  \centering
  \includegraphics[width=.9\linewidth]{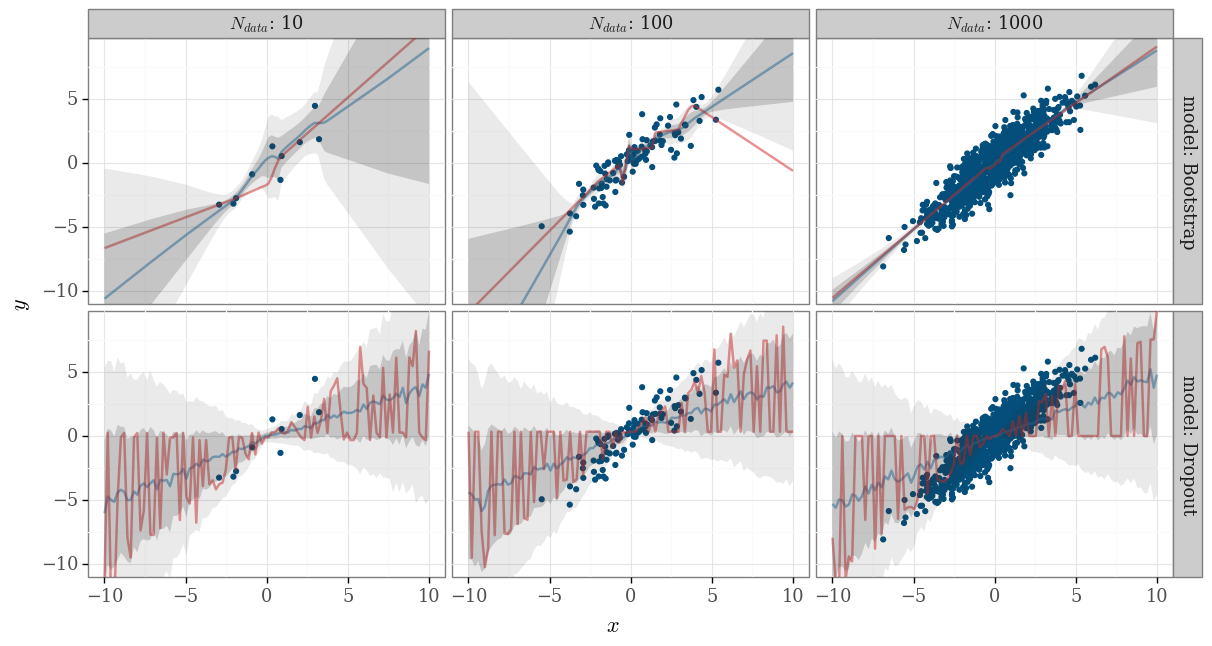}
  \caption{\small Dropout does not converge with increasing data even with a complex neural network. Grey regions indicate $\pm 1, 2$ standard deviations, the mean is shown in blue and a single posterior sample in red.}
  \label{fig:dropout_no_converge}
\end{figure}



\subsection{Variational inference and Bellman error}
\label{app:variational_inference}

Lemma \ref{thm:variational} highlights that the basic loss most commonly used by variational approximations to the value function are fundamentally ill-suited to the problem at hand \cite{lipton2016efficient,fortunato2017noisy}.
In Appendix \ref{app:experiment_details} we present results of such a variational approach, NoisyNet, to some of our benchmark reinforcement learning tasks.
As expected, the algorithm performs poorly even after extensive tuning.
At the heart of this issue is a sample-based loss that trains to match the \textit{expectation} of the target distribution, but does not attempt to match the higher moments of the uncertainty.
However, we could imagine an alternative approach that does aim to match the entire resultant distribution, for example via parameterized distribution and cross entropy loss; we leave this to future work.

Even where variational inference is employed correctly over mutliple timesteps, it may be difficult to encode useful prior knowledge in common variational methods.
First, many applications of variational inference (VI) model the distribution over network weights as a product of independent Gaussians \cite{blundell2015weight}.
These models facilitate efficient computation, but can underestimate the uncertainty and may be a poor choice for encoding prior knowledge.
Even if one were given a mapping of prior knowledge to weights, the confounding demands of good initialization for SGD training may interfere negatively \cite{glorot2010understanding}.
For this reason practical applications of VI to RL typically use very little prior effect, or even no prior regularization at all \cite{lipton2016efficient,fortunato2017noisy}.

\subsection{`Distributional reinforcement learning'}
\label{app:dist-rl}

Unlike the objections of Appendix \ref{app:variational_inference}, `distributional RL'\footnote{Any method for Bayesian RL might reasonably claim to be a distributional perspective on reinforcement learning. For this reason, we use quotation marks when we want to distinguish the specific form of distributional RL popularized by \cite{bellemare2016countbased}.} does learn a value function estimate through a distributional loss.
However, this distribution is a distribution over \textit{outcomes} and not a distribution over the \textit{epistemic uncertainty} in the mean beliefs.
This distinction between two types of uncertainty, (1.) things that you don't know and (2.) things that are stochastic, is a delicate one and is important to characterize correctly. Both are discussed under many names:
\begin{itemize}[noitemsep, nolistsep]
  \item[(1.)] `Reducible uncertainty' $\iff$ `epistemic uncertainty' $\iff$ `uncertainty',
  \item[(2.)] `Irreducible uncertainty' $\iff$ `aleatoric uncertainty' $\iff$ `risk'.
\end{itemize}
Typical decision problems may include elements of both types of uncertainty.
Flipping a coin we might want to know both (1.) our posterior beliefs over the probability of heads and (2.) a distribution that categorizes the likely possible outcomes.
However, it should be clear that the two concepts are fundamentally distinct.
For the purposes of exploration, the Bayesian uncertainty over (1.) should prioritize the acquisition of new knowledge.
`Distributional RL' approximates (2.) and its role is not exchangeable with (1.).

\begin{lemma}[Using `distributional RL' as a posterior can lead to arbitrarily bad decisions]
\label{thm:distributional}
\hspace{0.000001mm} \newline
Consider an agent with full information that decides between action $1$ with reward $\sim {\rm Ber}(0.5)$ and action $2$ with reward $1 - \epsilon$ for $0 < \epsilon < \frac{1}{2}$.
If the agent employs Thompson sampling correctly then it will pick action $2$ at every step with zero regret.
If the agent mistakenly employs Thompson sampling over its `distributional value function' then it will incur
$$ \Exp \left[ {\rm Regret} (T) \right] \ge \frac{1}{2} \left(\frac{1}{2} - \epsilon\right) T. $$
\end{lemma}

Lemma \ref{thm:distributional} shows that using the `distributional' value function approximating (2.) can be a poor proxy for the Bayesian uncertainty.
However, the Bayesian uncertainty can be a similarly poor proxy for the `distributional' value function.
This can be equally damaging, particularly if the agent has some some risk-sensitive utility with respect to cumulative rewards.
It is entirely possible to combine both notions of uncertainty in an agent, although for the goal of maximizing expected cumulative it is not entirely clear what is the benefit of modeling (2.).
Certainly, `distributional' agents have recently attained strong scores in Atari 2600 benchmarks but it is so far unclear exactly what the source of this outperformance comes from \cite{bellemare2017distributional,dabney2017qr}.
Possible explanations may include more stable gradients, bounded values and the `many predictions' hypothesis \cite{sutton2011horde}: that learning a distribution may effectively create a series of auxiliary losses.
We leave these questions for future work.

\subsection{Count-based uncertainty estimates}
\label{app:count}

Count-based approaches to exploration give a bonus to states that have not been visited frequently according to some density measure $p(x)$.
These methods have performed well in many sparse reward tasks such as Montezuma's Revenge, where visiting new states acts as a shaping reward for the true reward \cite{bellemare2016countbased}.
However, a count-based bonus is generally a poor approach to exploration beyond the tabular setting.
To see why this is the case note that the density measure of the states may not correlate well with the agent's uncertainty over the optimal policy in that state.
We can imagine situations both where the state is visually new, but an agent should still know exactly what to do; and also settings that are only delicately different to a common situation but still necessitate exploration of the optimal policy.

This disconnect shows up in problems as simple as the linear bandit.\footnote{Reward $r_t(x_t) = x_t^T \theta^* + \epsilon_t$ for some $\theta^* \in \Real^d$ and $\epsilon_t \sim N(0,1)$ \cite{RusmevichientongT2010}.}
Via a packing argument, an agent with count-based uncertainty will require $\tilde{O}\left(\frac{1}{\epsilon^d}\right)$ measurements to cover the space up to radius $\epsilon$.
By contrast, an agent that explores this space efficiently can resolve its uncertainty in only $\tilde{O}(d)$ measurements.
Thompson sampling with a linear model naturally recovers this performance \cite{Russo2014}.
The fact that this failure can arise even in a linear system, and even when the density can be estimated precisely, suggests that count-based exploration is not \textit{in general} an effective method for simultaneous exploration with generalization; even if it may be effective at some specific tasks.




\subsection{Ensembles without priors}
\label{app:no_prior}

This paper builds upon a line of research that uses an ensemble of trained models to approximate a posterior distribution.
Compared to previous works, our main contribution is to highlight the importance of a `prior' mechanism in ensemble uncertainty.
Figure \ref{fig:boot_vs_ensemble} presents an extremely simple example of 1D regression with a (20,20)-MLP and rectified linear units.
The data consists of $x_i = \frac{i - 5}{5}$ for $i=0,..,10$ and $y_i = 5 \Ind\{i=10\}$.

\begin{figure}[h!]
  \centering
  \includegraphics[width=\linewidth]{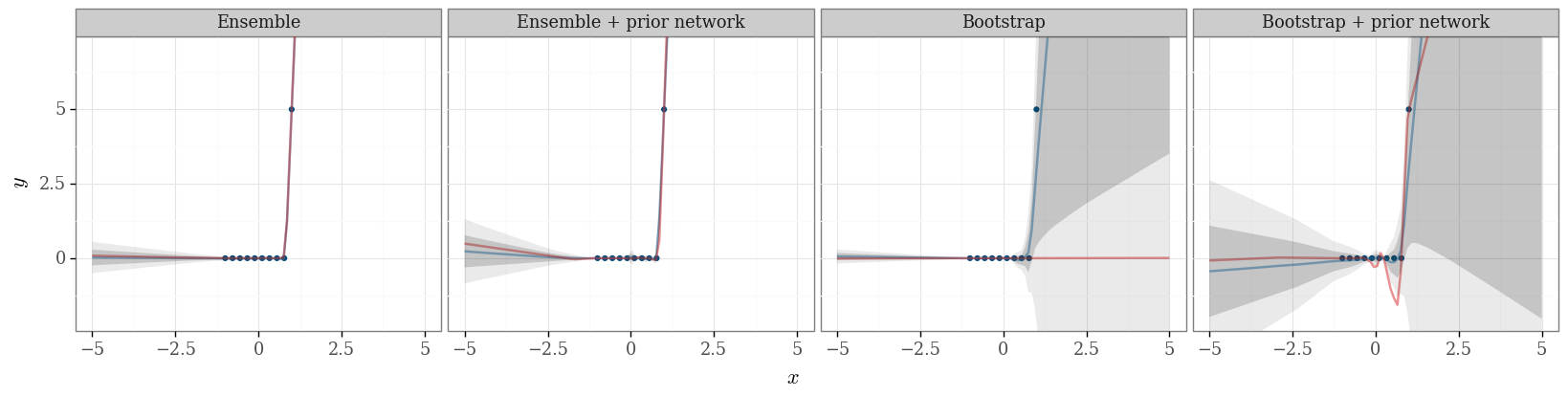}
  \caption{Posterior predictive distributions for ensemble uncertainty.}
  \label{fig:boot_vs_ensemble}
\end{figure}

The results above highlight the drawbacks of naive ensembles.
A pure ensemble trained from random initializations fits the data exactly and leads to almost zero uncertainty anywhere in the space \cite{lakshminarayanan2017simple}.
A bootstrapped ensemble takes the variability of the data into account and thus has a wide predictive uncertainty as $x$ grows large and positive.
However, where the data has target value zero, bootstrapping will always produce a zero target and consequently the ensemble has almost zero predictive uncertainty as $x$ becomes large and negative \cite{osband2016deep}.

This lack of prior uncertainty can lead to arbitrarily poor decisions, as outlined in \cite{osband2015bootstrapped}.
If an agent has only ever observed zero reward, then no amount of bootstrapping or ensembling will cause it to simulate positive rewards.
This issue is easily remedied by the addition of a prior mechanism, either through $l_2$ regularization to initial random weights \eqref{eq:l2_weight_blr}, or the addition of a fixed additive random `prior network' \eqref{eq:l2_weight_blr_2}.

\subsection{Summary}
We summarize the issues raised in Section \ref{sec:why_prior} in Table \ref{tab:effectiveness}.
This table is meant only as a rough summary and should not be taken as rigorous statement. Roughly speaking, a green tick means success, red cross means failure and a yellow circle means something in between.
{This paper proposes a combination of bootstrap sampling with prior function as an effective computational approximation to Bayesian inference in deep RL.}
Although our method is somewhat computationally expensive, since it requires training an ensemble of models instead of one, this computation can be done in parallel and so is amenable to large scale distributed computation.

\begin{table}[h!]
\centering
\caption{Important issues in posterior approximations for deep reinforcement learning.}
\vspace{2mm}
\label{tab:effectiveness}
{\small
\begin{tabulary}{\linewidth}{r|CCCCCC}
& Data conc. & Learned metric & Multi step & Works in noise & Prior effect & Cheap compute  \\
\hline
Dropout \cite{Gal2016Dropout} & \cross & \tick & \cross & \ok & \cross & \tick \\
NoisyNet \cite{fortunato2017noisy}& \ok & \tick & \cross & \tick & \cross & \tick \\
BBB / VI \cite{blundell2015weight} & \ok & \tick & \cross & \tick & \ok & \tick \\
Density count \cite{bellemare2016countbased} & \tick & \cross & \tick & \cross & \ok & \tick \\
`Distributional' RL \cite{bellemare2017distributional} & \cross & \ok & \tick & \ok & \cross & \tick \\
Ensemble \cite{lakshminarayanan2017simple} & \cross & \tick & \tick & \cross & \cross & \cross \\
Bootstrap \cite{osband2016deep} & \tick & \tick & \tick & \tick & \cross & \cross \\
{\bf Bootstrap + prior }& \tick & \tick & \tick & \tick & \tick & \cross \\
Exact Bayes& \tick & \tick & \tick & \tick & \tick & \cross \cross \cross \\
\end{tabulary}
}
\end{table}

\newpage

\end{document}